# Scan-to-BIM for As-built Roads: Automatic Road Digital Twinning from Semantically Labeled Point Cloud Data

Yuexiong Ding[a,b,c], Mengtian Yin[c], Ran Wei[c], Ioannis Brilakis[c], Muyang Liu[a,b], Xiaowei Luo[a,b*]

[a]Department of Architecture and Civil Engineering, City University of Hong Kong, Hong Kong, China

[b]Architecture and Civil Engineering Research Center, Shenzhen Research Institute of City University of Hong Kong, Shenzhen, China

[c]Department of Engineering, University of Cambridge, Civil Engineering Building, JJ Thomson Avenue 7a, Cambridge CB3 0FA, UK

## Abstract

Creating geometric digital twins (gDT) for as-built roads still faces many challenges, such as low automation level and accuracy, limited asset types and shapes, and reliance on engineering experience. A novel scan-to-building information modeling (scan-to-BIM) framework is proposed for automatic road gDT creation based on semantically labeled point cloud data (PCD), which considers six asset types: **Road Surface**, **Road Side** (Slope), **Road Lane** (Marking), **Road Sign**, **Road Light**, and **Guardrail**. The framework first segments the semantic PCD into spatially independent instances or parts, then extracts the sectional polygon contours as their representative geometric information, stored in JavaScript Object Notation (JSON) files using a new data structure. Primitive gDTs are finally created from JSON files using corresponding conversion algorithms. The proposed method achieves an average distance error of 1.46 centimeters and a processing speed of 6.29 meters per second on six real-world road segments with a total length of 1,200 meters.

**Keywords:** Geometric digital twin; Point cloud data; As-built road; Scan-to-building information modeling; Sectional polygon contour; Geometric information extraction;

## 1. Introduction

The current socio-economic development and people's quality of life are highly dependent on roads, especially the coverage of highways. An unofficial statistic shows that the global road mileage has exceeded 63 million kilometers [1]. According to the United Kingdom (U.K.) Department for Transport, approximately 86% of passengers and 81% of goods were transported by road in 2022 [2]. Though motorways and "A" roads in the U.K. only account for 13% of the national road network by length, 65% of motor vehicle miles traveled were on these two types of road [3]. National roads will undoubtedly expand with the socio-economic development, making maintenance, renovation, and expansion work increasingly complex and challenging, especially for those early-built roads. The road digital twin (RDT) has now become one of the promising approaches for the current situation, as it can provide integrated, structured, and managed data for everyday operations and management works in a unified manner. The promotion and implementation of RDT can help improve multiple aspects of social life, such as improving the quality of public services [4], enhancing national productivity [5], and promoting a net-zero sustainable circular economy [6].

Generally, a digital twin (DT) can be considered composed of a three-dimensional (3D) digital model/portrait and all kinds of current and historical information generated from it [7]. Within the built environment, a DT is a realistic digital representation of assets, processes, and systems [8]. Therefore, the first step of RDT implementation is constructing a 3D geometric model for the targeted road, known as the geometric digital twin (gDT) [9]. However, many roads were built without digital drawings [10], which makes it challenging for automatic gDT creation, leaving only the manual creation method that requires much labor. Though some roads retain construction drawings, such as paper or PDF files, they are unstructured and provide very limited substantial help for the automatic gDT creation. Furthermore, relevant departments are unwilling to promote digital twinning for these early-built roads because they believe that the cost of manual gDT creation currently exceeds the future benefits.

Many explorations have been conducted to pursue automatic gDT creation for as-built infrastructures. For example, many commercial building information modeling (BIM) software suppliers, such as Autodesk, Bentley, Trimble, AVEVA, and ClearEdge3D, have developed various state-of-the-art solutions/tools/plugins based on light detection and ranging (LiDAR) point cloud data (PCD) [9,11,12] since LiDAR PCD has the highest 3D spatial perception accuracy. However, they can only fit limited objects with standardized pre-defined shapes and still require a lot of customized editing/adjusting operations [11,13]. Though many relevant studies have also been conducted in academia, they still have many distinct limitations, such as low automation level [14–16] and accuracy [9,16], limited types and shapes of road assets [14–16], and reliance on engineering experience/specifications [15,16]. Among them, the work conducted by Lu and Brilakis [9] should be the state-of-the-art practice, which achieved automated gDT creation of four reinforced concrete bridge components from semantically labeled PCD, with an average distance error of 7.05 centimeters (cm). In addition, almost all relevant studies use specific 3D/BIM file formats (e.g., the Industry Foundation Classes (IFC)) to store the created gDTs, resulting in larger files and inefficient data input/output (I/O), thus hindering efficient data transmission and parsing between the client and server in the Internet of Things (IoT) era.

This paper presents a novel framework for the automatic gDT creation of multiple road assets based on the geometric information extracted from semantically labeled LiDAR PCD. There are mainly three challenges when implementing the proposed framework: 1) how to extract geometric information of road assets with arbitrary shapes given a typical road containing multiple shaped components/assets; 2) is there a more efficient strategy to structurally store the extracted geometric information of different road assets; and 3) how to create gDTs for different road assets from the stored geometric information. Therefore, the proposed framework contains a series of targeted solutions/algorithms to address the mentioned challenges. Briefly, the framework first applies clustering and part segmentation to segment the PCD with the same semantic label into multiple spatially independent instances or parts. Targeted algorithms are then developed for different asset instances or parts to extract the sectional polygon contours as their representative geometric information (challenge 1), which is structurally stored in JavaScript Object Notation (JSON) files using a new compressed data structure (challenge 2). Finally, corresponding conversion algorithms are developed to construct primitive gDT representations from extracted/saved geometric information (challenge 3). The framework is verified on six real-world road segments with a total length of 1,200 meters (m), which achieves an average distance error of 1.46cm with an average processing speed of 6.29 meters per second (m/s).

The novelties and field contributions can be summarized as follows: 1) This study proposes and implements a novel framework containing a series of solutions/algorithms for full process automation from semantically labeled PCD to gDT outputs; 2) The proposed framework unifies the geometric information extraction of diverse road assets into the extraction of sectional polygon-contours, theoretically applicable to road assets of arbitrary shapes; 3) Unlike other studies saving the 3D/BIM file format, this study permanently stores the original extracted geometric information using a new structured and compressed data format to achieve more efficient storage and higher scalability; 4) Compared to existing studies and manual gDT creation methods, the proposed framework achieves the best results both in accuracy and efficiency; 5) The automatic framework implemented in this study is expected to save considerable cost of labor, time, and money for twinning as-built roads in the future.

The rest of the paper is organized as follows: Section 2 reviews various related works to provide background information and the latest research progress. Section 3 elaborates on the proposed framework. Section 4 conducts real-world experiments and evaluations, while Section 5 makes further discussions. Section 6 concludes the study, as well as the limitations and future works.

## 2. Literature review

### 2.1 DT in construction

There are various definitions for digital twins (DT) within the industry and academia. Fundamentally, DT can be defined as the constantly evolving digital contours of a physical object's historical and current behaviors [7]. In other words, a typical DT comprises a 3D geometric model with geometric dimensions and tolerances and all current and historical data generated from it.

DT and relevant technologies have been widely applied in many construction-related fields [17,18]. Research has shown that DT can improve productivity and help simplify predictive maintenance [19], especially in the digital operation and maintenance of as-built buildings or infrastructure [20]. For example, Peng et al. [21] constructed a hospital DT integrating building information modeling (BIM), internet of things (IoT), and mixed reality (MR) technologies, which can accurately control the operation of facilities and reduce the frequency of equipment maintenance and failures. Based on various finite element models, Lin et al. [14] proposed a DT-based collapse fragility assessment method for long-span cable-stayed bridges under strong earthquakes. Hosamo et al. [22] developed a DT-based predictive maintenance framework for air handling units based on BIM, IoT, and semantic technologies to automatically detect and diagnose faults in these facilities. Gao et al. [23] proposed a DT communication framework informed by Artificial Intelligence of Things (AIoT) to support intelligent bridge operation and maintenance with high efficiency, low latency, and excellent fault tolerance.

Though studies mentioned above have extensively demonstrated the effectiveness and positive feedback of DT in many construction-related areas, they just used existing or manually created gDTs and did not focus on relevant ideas or solutions for automatically performing digital twinning on as-built buildings/infrastructures.

## 2.2 Twinning the built environment

To create a gDT for an existing physical object, it is usually necessary to use some devices to collect relevant information about the target object in the physical world. According to the data sources, the explorations of twining the built environment can be divided into image-based, reconstructed PCD-based, and LiDAR PCD-based methods.

*1) Image-based methods*

Image-based methods have obvious advantages, such as simple, fast, low-cost, etc. The input of these methods can be images, 2D drawings, and other 2D data. For example, Lu et al. [24] developed a fuzzy-based system to construct as-is IFC BIM objects driven by images. Furthermore, they proposed a semi-automatic geometric digital twinning framework for existing buildings based on images and 2D computer-aided design (CAD) drawings [25]. However, the construction from 2D images to 3D geometry is an ill-posed problem as the dimensional information provided by 2D image data is limited. Therefore, these methods require a significant amount of manual intervention or prior experience as additional information input to correct ill-posed biases.

*2) Reconstructed PCD-based methods*

Compared to image data, PCD provides three-dimensional information about the physical object, effectively solving the ill-posed problem of 2D data to 3D geometry. Currently, there are two main methods for capturing PCD: photogrammetry and laser scanning or light detection and ranging (LiDAR). Photogrammetry can be seen as an advanced image-based method that reconstructs the PCD of the target scene/object from multiple 2D images in different capture views using structure from motion (SFM) or multi-view stereo (MVS) technologies. For example, Pan et al. [26] proposed a semi-automated framework for generating the structural surface models of heritage bridges. This framework first used the SFM to reconstruct the PCD of bridges from a series of images and then applied the Poisson surface reconstruction algorithm to generate structural surfaces for each segmented structural element. Similarly, Hu et al. [27] developed an encoder-decoder network using both multi-view images and a photogrammetric PCD as input to model a high-level structural relation graph and low-level 3D geometric shapes. The Poisson surface reconstruction based on the decoder output showed better results than that on manually segmented dense PCD. Ma et al. [28] developed a measurement system to get triangle meshes of sewer pipelines' potholes from the reconstructed dense PCD based on SFM and MVS. Jiang et al. [16] used various aerial photogrammetry mapping data, such as aerial photography, digital surface model (DSM), and topographies, to construct digital twins of existing highways based on engineering expertise semi-automatically, with a distance error of 15.76cm. However, photogrammetry is highly sensitive to shadow and lighting conditions, generating unstable and lower accuracy/quality PCD in outdoor conditions.

*3) LiDAR PCD-based methods*

Light detection and ranging (LiDAR), also called laser imaging detection and ranging, is a remote sensing method that irradiates objects using near-infrared, visible, or ultraviolet light and then detects the reflected light with an optical sensor to measure the distance. LiDAR is characterized by being able to detect not only accurately the distance to the target object but also its 3D position and shape.

Given the higher accuracy of 3D shape information captured by the LiDAR PCD, some commercial BIM software providers have developed various plugins/tools to enable manually or automatically place/create gDTs based on the scanned PCD, which, to some extent, speeds up the manual work of twinning the built environment. Major suppliers such as Autodesk,

Bentley, Trimble, AVEVA, and ClearEdge3D offer the most advanced solutions based on LiDAR PCD [9,11]. However, these tools can only fit PCD with standardized pre-defined shapes (such as rectangular walls, pipes, valves, flanges, and steel beams) instead of PCD with arbitrary shapes [12]. Some provide (e.g., Revit of Autodesk) further flexibility, allowing users to freely edit the pre-defined family shapes to fit the target PCD more accurately. However, it still requires a lot of manual editing/adjusting operations after placing them in the right places. Relevant statistics show that 95% of the total modeling time is spent on customizing shapes and fitting them to point clusters [13], and the sidewalks and road surface boundary adjustments are the most time-costly [11] in road twinning cases.

Many academic studies also try to achieve automatic digital twinning for as-built infrastructure (e.g., bridges, roads). For example, Lu and Brilakis [9] developed a slicing-based fitting method to generate gDTs for four components (slab, pier, pier cap, and girder) of existing reinforced concrete bridges from labeled PCD, achieving an average modeling distance of 7.05 centimeters. Soilán et al. [14] proposed a semi-automated framework to generate an IFC-compliant file from the PCD of highways. However, only the alignments and centrelines of road surfaces/lanes were extracted and represented. Similarly, Justo et al. [15] semi-automatically generated IFC-compliant representations for road surfaces, traffic signs, and guardrails. However, they extracted limited information about these assets, such as the road centerline and guardrail position, and then generated and placed IFC entities according to given engineering specifications (UNE 135121:2012). Besides, the diversity of traffic signs considered is very limited in Justo's research, with only one traffic sign being studied. Wang et al. [29] conducted semantic segmentation for the PCD to recognize road surfaces and then extracted geometric information from the segmented road surface points. The digital road models were finally created manually based on the extracted geometric information via Dynamo and Revit.

### 2.3 Research gaps

According to the literature review, existing studies still have the following limitations: 1) Low level of automation. Most existing studies were semi-automated and needed a large amount of manual intervention when extracting target objects' geometric features or creating gDT representations, resulting in low twinning efficiency. 2) Low accuracy. Though some studies have achieved centimeter-level distance error, there is still significant room for improvement; 3) Limited asset types and shapes. Most existing studies only focused on extracting geometric information of road surfaces and lane markings with regular shapes but less on road signs/lights and guardrails with arbitrary shapes. 4) Overreliance on engineering experience. Due to limited or low-quality information provided by the input data, some studies had to rely on specified engineering experience or specifications, which only generated textbook-style roads rather than exact twins of existing roads in the real world.

## 3. Methodology

This study focuses on automatically creating gDTs for multiple road assets from road PCD with semantic labels. Therefore, any issues related to the raw PCD, such as the acquisition and quality of the raw PCD and the accuracy of its semantic labels, are beyond this study's scope.

Figure 1 shows the overall research framework of the proposed method. Based on the occurrence percentage statistics of highway objects from [11], six types of road assets are considered: Road Surface, Road Side (Slope), Road Lane (Marking), Road Sign, Road Light, and Guardrail, which are further classified into Plane-like (Road Surface/Side/Lane), Pole-like (Road Sign/Light), and Guardrail hyper-assets. First, a density-based clustering method is applied to segment each PCD with the same semantic label into multiple spatially independent instances. Pole-like asset instances are further segmented into numerous functional parts using a part segmentation model. Targeted algorithms are then developed for each hyper-asset category or functional part to extract 3D geometric information from PCD instances. Unlike other existing methods, the extracted geometric information is not directly converted into a specific high-level 3D file format but is formatted using a novel structured and compressed method and saved in JavaScript Object Notation (JSON) files for permanent storage. Finally, several corresponding conversion algorithms are developed to create 3D gDTs, which are low-level primitive geometrical representations consisting of a set of point, line, and face lists, making it easy to convert to various 3D/BIM file formats (e.g., IFC, PLY, OBJ, STL, etc.). This study adopts the IFC file format as an example to represent and visualize the created gDTs.

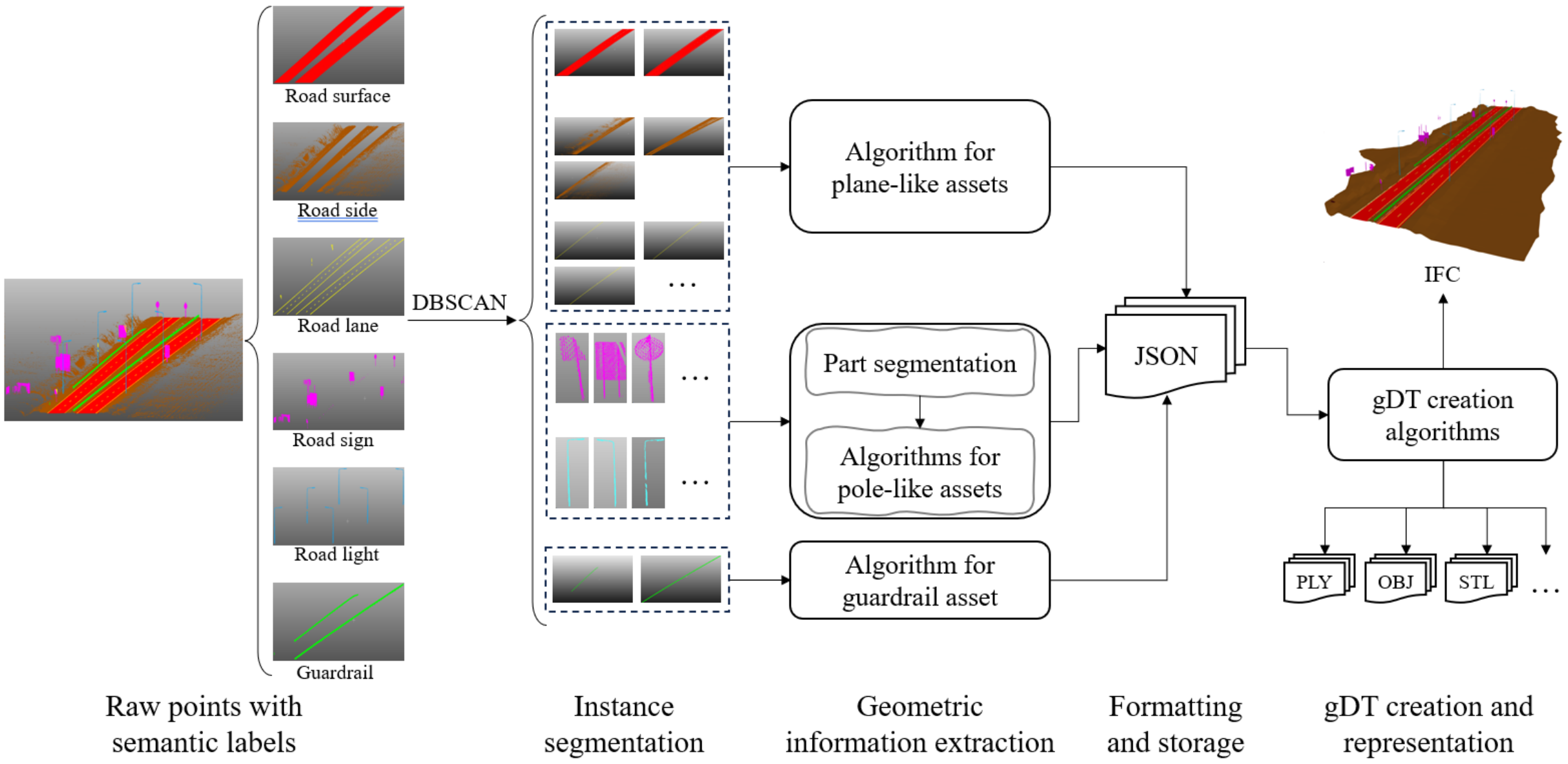

Figure 1. Proposed research framework.

### 3.1 Instance segmentation

As shown in Figure 1, raw PCD with the same semantic label still contains multiple individual asset instances. Therefore, instance segmentation is first performed on the raw PCD to facilitate the instance-level geometric information extraction. Different asset instances are spatially independent and maintain a certain distance from each other within a PCD with the same semantic label, i.e., points are dense if there is an instance, while points in the interval between instances are sparse, as shown in Figure 2 a). Based on this characteristic, a typical density-based spatial clustering algorithm, density-based spatial clustering of applications with noise (DBSCAN), was applied in this study for fast instance segmentation within a PCD with the same semantic label. The significant advantage of the DBSCAN is its fast clustering speed and ability to effectively handle noise points and discover spatial clusters of arbitrary shapes.

The main idea of DBSCAN consists of two parameters (Eps and MinPts) and three fundamental concepts (directly density-reachable, density-reachable, and density-connected). The parameter Eps ($\varepsilon$) is the radius of the neighborhood when defining density, and MinPts is the minimum neighborhood number when defining core points. Based on these two parameters, for any point $x \in D$, its $\varepsilon$-neighborhood is defined as $N_{\varepsilon}(x) = \{y \in D | dist(x, y) \leq \varepsilon\}$, and $|N_{\varepsilon}(x)|$ is the density of point $x$. According to the density definition, points can be further divided into core, border, and noise points, as shown by red, blue, and gray color points in Figure 2 b). If $|N_{\varepsilon}(x)| \geq MinPts$, then $x$ is a core point. The set of core points is defined as $D_c = \{x \in D||N_{\varepsilon}(x)| \geq MinPts\}$, while the set of non-core points is $D_{nc} = \{x \in D||N_{\varepsilon}(x)| < MinPts\}$. If $x \in D_{nc}$, and there is a core point in its neighborhood, then $x$ is a border point, i.e., $D_{bd} = \{x \in D_{nc} | dist(x, y) \leq \varepsilon, y \in D_c\}$. Points in $D$ that neither belong to $D_c$ nor belong to $D_{bd}$ are noise points, i.e., $D_{noi} = D - (D_c + D_{bd})$. Let $x, y \in D$, if $y \in N_{\varepsilon}(x)$, then $y$ is directly density-reachable from $x$. Assuming that there is a sequence of points $x_1, x_2, \dots, x_n$, if $x_{i+1}$ is directly density-reachable from $x_i$, then $x_n$ is density-reachable from $x_1$. Density reachability does not have symmetry. For $x, y, z \in D$, if $y$ and $z$ are both density-reachable from $x$, then $y$ and $z$ are density-connected. It is evident that density connectivity has symmetry. After determining the density reachability and connectivity between points, DBSCAN defines clusters as the maximum set of density-connected points. A non-empty subset $C$ is a cluster of $D$ if $C$ satisfies: 1) for $x, y \in D$, if $x \in C$ and $y$ is directly density-reachable from $x$, then $y \in C$; 2) if $x \in C$ and $y \in C$, then $x$ and $y$ are density-connected.

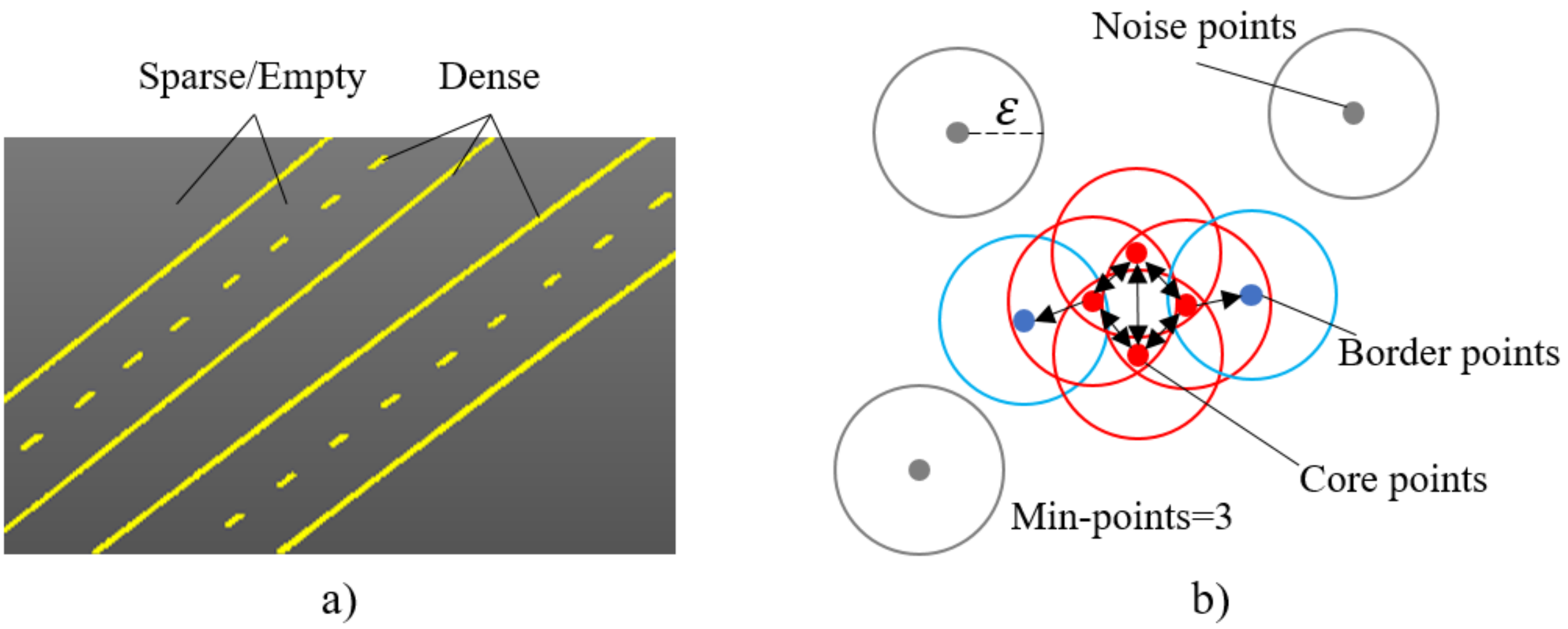

Figure 2. a) Spatial distribution characteristics of the same semantic points. b) DBSCAN clustering algorithms.

### 3.2 Geometric information extraction

Though instance segmentation locates the operation object to a single instance, instances of different assets have diverse shape contours. For example, Road Surface, Road Side, and Road Lane instances are distributed in a plane-like pattern, while others are distributed in 3D modes. In addition, the shapes of different parts within the same instance are also not similar. For example, a Road Sign/Road Light instance usually consists of one or two poles and one or more functional heads (e.g., panels and lights/lamps). Therefore, this study further classified six considered assets into three hyper-asset categories: Plane-like, Pole-like, and Guardrail assets, then developed targeted geometric information extraction algorithms for each hyper-asset. For Pole-like assets, a deep learning-based part segmentation model was applied to segment the instance into more specific functional parts, narrowing the minimal extracting unit into the part level.

#### 3.2.1 Basic information

This section briefly introduces some essential concepts/tools/methods used in this study to facilitate understanding subsequent algorithms.

1) ***Polygon.*** This study mainly used polygons to represent the extracted geometric contours of the instance/part. As shown in Figure 3 a), a typical polygon consists of a closed external polyline (shell) and zero or multiple closed internal polylines (holes). A polyline comprises a set of points/vertices and line segments. The polyline with the same starting and ending point is the closed/loop polyline. For more information about polygons, please refer to [30]. The Shapelysmooth tool [31] provides multiple algorithms (e.g., Taubin, Chaikin) for polygon/polyline smoothing.

2) ***Alphashape.*** The Alphashape, known as the rolling-ball method, is an algorithm that extracts boundary polygons containing a set of 2D points. As shown in Figure 3 b), the Alphashape algorithm is like rolling a ball with a radius of $1/\alpha$ on a set of 2D points. The contour polygon of the point set can be obtained by connecting the rolling paths of the ball. Setting different $\alpha$ value can control the refinement level of the extracted polygon: small $\alpha$ for coarse polygon (may not have holes) and large $\alpha$ for fine polygon (may contain many holes). For more information about the Alphashape algorithm, please refer to [32].

3) ***Voronoi diagram-based polygon centerline extraction.*** A Voronoi diagram is a partition of a plane into regions close to each of a given set of seed points/objects. In the Voronoi diagram, there is a corresponding region for each seed point, called a Voronoi cell, also known as the Thiessen polygon. The distance from any point within a Thiessen polygon to the seed point that makes up the polygon is smaller than the distance to other seed points. Therefore, the Voronoi diagram is an equidistant subdivision of a plane area and can be used to extract a polygon’s centerline. As shown in Figure 3 c), let the polygon’s vertices be the seed points to construct a Voronoi diagram. Then, the polygon centerline consists of line segments inside the polygon. For more information regarding the polygon centerline extraction, please refer to [33].

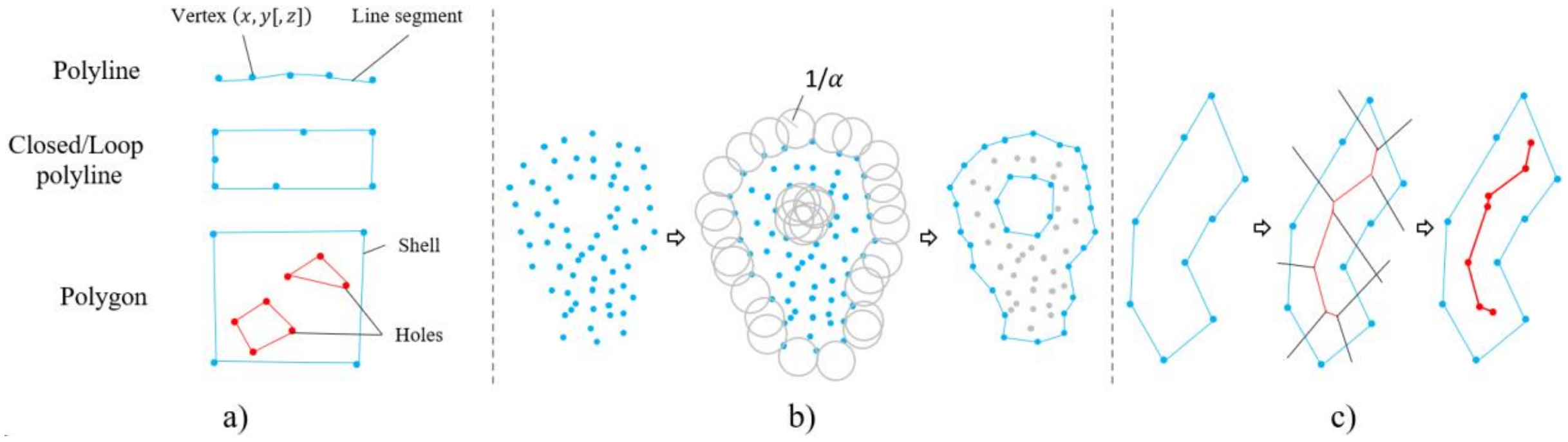

Figure 3. Demonstrations of a) polyline and polygon, b) Alphashape algorithm, and c) Voronoi diagram-based polygon centerline extraction.

4) ***Centreline-based polygon split.*** This method mainly splits a polygon based on the extracted centerline(s). The general steps are shown in Figure 4 a). First, each centerline should be resampled using the polyline interpolation algorithm [30] to obtain equidistant line segments. For each line segment of the new centerline, create its perpendicular bisector and extend it until both ends intersect with the polygon. The polygon is finally divided by these perpendicular bisectors.

5) ***Circle-K-neighbors approximation (V1 and V2).*** This method is used to convert 2D points to 3D points given a reference 3D point set. V1 is for point-to-point conversion, while V2 is for point-to-point-pair conversion. As shown in Figure 4 b), the V1 method is used to approximate a 2D point in the X-Y plane. First, search reference points ($N$) within the radius $r$ of the target 2D point in the 2D plane. If there is no reference point within the given radius, then use the $K$ closest reference points to the target point. In the V1 case, the third-axis value of the target point is the average of the third-axis values of the searched reference points (e.g., $[x', y']$ to $\left[x', y', Avg\left(z_1, \dots, z_{N(K)}\right)\right]$). In the V2 case, a point-pair is generated using the averages of the top-$k$ ($2k \leq N(K)$) maximum and minimum of the searched reference points (e.g., $[x', y']$ to $[x', y', Avg(z_1^{max}, \dots, z_k^{max})]$ and $\left[x', y', Avg(z_1^{min}, \dots, z_k^{min})\right]$).

6) ***Part segmentation.*** This method is used to decompose Pole-like asset instances into specific functional parts. This study adopted the Self-positioning Point-based Transformer (SPoTr) [34] as the part segmentation model and retrained it on the PolePartSeg dataset [35]. The retrained SPoTr model achieved 91.86% mean Intersection over Union (mIoU) on PolePartSeg's test set. The PolePartSeg dataset identified 12 part categories (Pole, Street Light, Traffic Light, Traffic Sign, Street Sign, Direction Sign, Flag/Banner, Decoration/Flowers, Horizontal Bar, Camera, Public Transport Sign, Bollard). This study only used six categories of PolePartSeg and reclassified them into **Pole**, **Light** (Street Light), and **Panel** (Traffic Sign, Street Sign, Direction Sign, and Public Transport Sign), as shown in Figure 4 c). Since this study mainly focuses on extracting geometric information from labeled PCD, please refer to [34] and [35] for more details about the modeling of the part segmentation model.

7) ***Coordinate system.*** The right-hand coordinate system was adopted, in which the X, Y, and Z axis are rightward, forward, and upward, respectively. The X-Y plane refers to the ground plane.

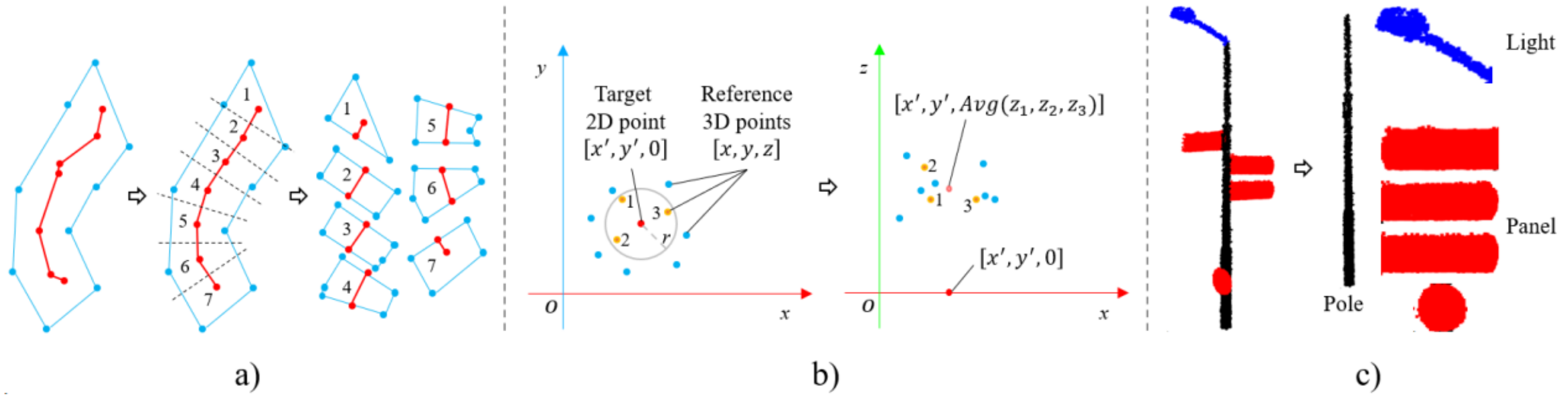

Figure 4. Demonstrations of a) centerline-based polygon segmentation, b) the Circle-K-neighbors approximation V1 for 2D-3D point conversion, and c) part segmentation for pole-like instances.

### 3.2.2 Plane-like asset

Plane-like assets include Road Surface, Road Side, and Road Lane assets. Figure 5 demonstrates the algorithm's overall procedures for extracting geometric information of Plane-like asset instances. Specifically, the algorithm can be described as the following steps:

- Step 1: Data preprocessing. This step is optional and mainly includes two operations: downsampling and removing outliers. Downsampling is used to reduce the point density of the instance and accelerate the processing speed of subsequent steps. Outlier removal can further exclude misclassified points in semantic/instance segmentation.

- Step 2: Project points onto the X-Y plane. This is because the geometric contour information of Plane-like asset instances is only presented on the X-Y plane, and the Z-axis direction only contains elevation information of the instance. Projecting 3D points onto the 2D plane once again accelerates the processing speed. Simply perform $[x, y, z] \rightarrow [x, y]$ to complete the fast projection operation.

- Step 3: Extract the fine (Step 3-1) and coarse (Step 3-2) polygons of the projected 2D points using the Alphashape. A fine polygon includes a shell polyline and several internal hole polylines, while a coarse polygon only has the shell polyline. This study applied $\alpha = 0.1$ and $\alpha = 10$ for extracting the fine and coarse polygons.

- Step 4: Extract the centerline(s) of the coarse polygon. For each extracted centerline, execute Step 5.

- Step 5: Split the coarse polygon using the centerline. For each sub-polygon, execute Step 6.

- Step 6: Perform grid-partition for the sub-polygon. First, rotate the sub-polygon around its center point ($p_c^i = [x_c^i, y_c^i]$) by $-\theta_i$, where $\theta_i$ is the included angle between its centerline segment and the X-axis. Then, create multiple grid polygons using the unit size of $grid_width \times grid_length$ based on the maximum and minimum values of the sub-polygon. Finally, rotate all the grid polygons around the center point ($p_c^i$) by $\theta_i$ to obtain the grid-partitioning results of the sub-polygon. The grid unit size determines the density of subsequent elevation sampling: small grids lead to high-density elevation sampling, while large grids result in low-density elevation sampling.

- Step 7 and Step 8: Merge all grid polygons.
- Step 9: Intersect the fine polygon with all grid polygons. This step mainly removes the point-free areas in the coarse polygon, i.e., excluding the hole areas.
- Step 10: Convert 2D polygons to 3D. The Circle-K-neighbors approximation V1 method is applied in this step to convert all 2D points of the obtained polygons into 3D points, using the reference 3D points obtained in Step 1.

Note that Road Lane asset instances do not need grid partition since they are relatively small. This study only executed Steps 1, 2, 3-1, and 10 to extract geometric information of Road Lane instances.

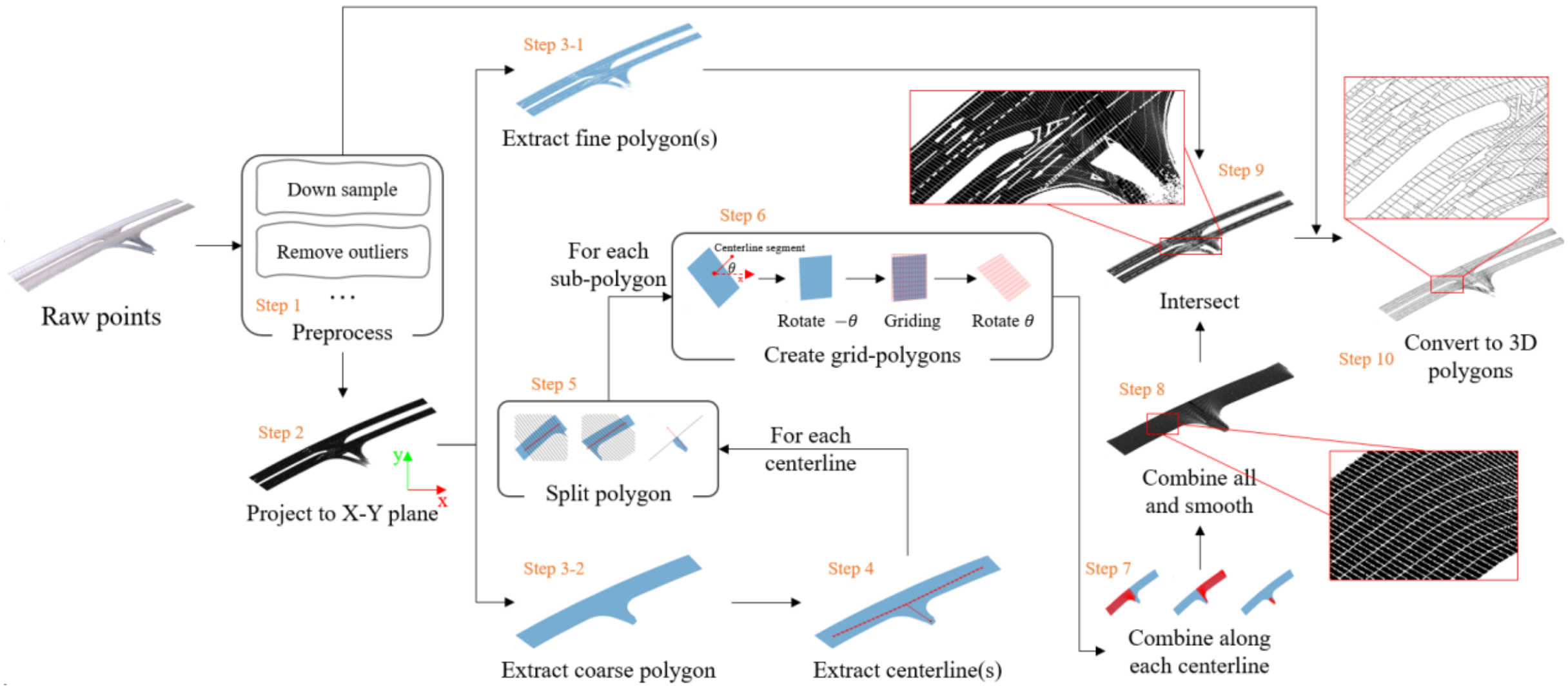

Figure 5. Geometric information extraction algorithm for Plane-like asset instances.

### 3.2.3 Guardrail asset

The Guardrail assets considered in this study mainly consist of two types: "T" and "#" shaped guardrails. Unlike Plan-like assets, Guardrail assets have geometric contours in both the X-Y and X-Z (or Y-Z) planes. Therefore, two "projection (3D-2D), inverse projection (2D-3D)" operations are applied to extract the 3D geometric information of Guardrail asset instances. The algorithm procedures are shown in Figure 6, which the following steps can describe:

- Steps 1 and 2: Data preprocessing, then projecting onto the X-Y plane.
- Step 3: Extract the polygonal contour of the projected points in the X-Y plane ($\alpha = 1.0$), then extract the centerline of the polygon. For each centerline, execute Step 4.
- Step 4: Split the polygon in the X-Y plane based on the centerline. For each sub-polygon, execute Step 5.
- Step 5: 3D Point block rotation. First, find all preprocessed 3D points contained in the sub-polygon, i.e., 3D points that satisfy "$[x, y]$ in the sub-polygon". Then, rotate these 3D points around the Z-axis by $-\theta_i$, with the center point of the sub-polygon ($p_c^i = [x_c^i, y_c^i]$). $\theta_i$ is the included angle between the sub-polygon's centerline segment and the X-axis.
- Step 6: Create a new straight guardrail PCD. Guardrails sometimes might be curved, making obtaining geometric descriptors for curved parts difficult. Besides, there may be an overlap between two adjacent 3D point blocks after the rotation in Step 5. Therefore, a new straight guardrail PCD is created in this step by combining (Step 6-1) and offsetting (Step 6-2) all rotated 3D point blocks belonging to the centerline.

- Step 7: Project the new guardrail PCD onto the X-Z plane (Step 7-1) and extract its polygonal contour ($\alpha = 10$) in the X-Z plane (Step 7-2).
- Step 8: Use the Circle-K-neighbors approximation V2 to convert 2D polygons into 3D polygon pairs.
- Step 9: Perform the corresponding inverse operations done in Step 6 on the 3D polygon pairs. First, segment the 3D polygon pair based on the starting positions of shifted point blocks in Step 6 (Step 9-1). Then, inverse the movement and rotation for each sub-polygon pair based on metadata saved in Step 6 (Step 9-2).
- Step 10: Combine all polygon pairs of each centerline.

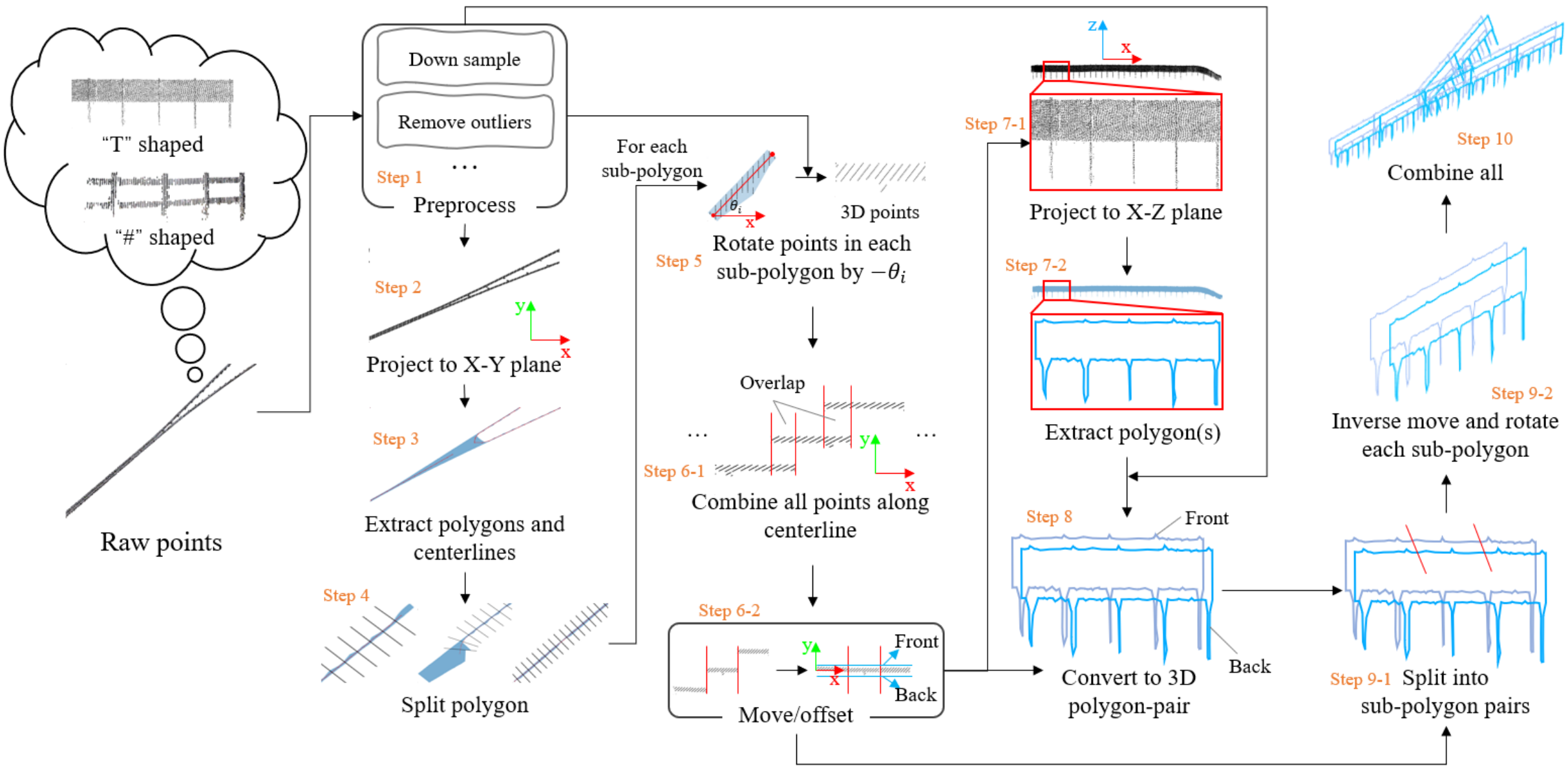

Figure 6. Geometric information extraction algorithm for Guardrail asset instances.

### 3.2.4 Pole-like asset

Plane-like assets include Road Sign and Road Light assets. Unlike Plane-like and Guardrail assets consisting of single shapes, Pole-like assets are typically composed of various geometric shapes. A typical Pole-like instance usually consists of one or two (or more) poles and one or more functional heads (e.g., panels and lights/lamps). Therefore, for each Pole-like instance, the part segmentation method is first applied to segment the instance into more specific parts of Pole, Light, and Panel types. Since some pole-like instances have multiple poles/lights/panels, the DBSCAN method is then applied to further segment the point set with the same part label into individual parts. Three targeted algorithms are finally designed to extract geometric information from each single part.

*1) Pole part*

Due to the limited precision of the raw PCD, the edge information of pole objects has been lost. For simplicity, this study treated all pole objects uniformly as cylindrical shapes. In addition, the size/diameter of some poles (e.g., poles of the road lights) usually varies with their height. Therefore, fitting a fixed-diameter cylinder is also not an ideal solution. In fact, the valuable geometric feature/descriptor of a cylinder at a certain height is its section 2D circle at this height. Therefore, the geometric information extraction algorithm for pole objects can be summarized as equally dividing the pole instance along the Z-axis and fitting the section circle in the X-Y plane. As shown in Figure 7 a), the algorithm can be described by the following steps:

- Step 1: Divide the pole equally along the z-axis with a fixed height of $\Delta h$. $\Delta h$ can be considered as the sampling frequency or granularity. The smaller the $\Delta h$, the more cross-sectional samples are collected, thus obtaining more detailed geometric contour information. For each divided sub-part, execute Step 2.

- Step 2: Extract the approximate circular polygon of the sub-part in the X-Y plane. First, project all points of the sub-part onto the X-Y plane (Step 2-1). Then, fit the smallest circle containing all projected points (Step 2-2). Third, draw $N$ rays with angles in an arithmetic sequence from the circle center. Take the intersection point between the ray with zero-angle and the circle as the starting point of the polygon, and order other points counterclockwise in sequence (Step 2-3). Finally, convert the 2D polygon to 3D by simply using the average Z-axis values of all points in the sub-part (Step2-4).

- Step 3: Collect all 3D polygons in sequence.

*2) Panel part*

Panel objects can be regarded as two faces (or polygons) with the same shape separated by a certain distance (the thickness of the panel). Therefore, the key to the algorithm for panel objects lies in extracting their front and back face polygons. Figure 7 b) demonstrates the extracting procedures, mainly including the following steps:

- Step 1: Rotate the panel to parallel the X-axis. Specifically, project the panel points onto the X-Y plane (Step 1-1). Then, use the least squares method to fit a straight line in the X-Y plane to obtain the angle ($\theta$) between the panel and the X-axis (Step 1-2). Finally, rotate the 3D panel points around the Z-axis by $-\theta$ (Step 1-3).
- Step 2: Extract the panel's face polygon. First, project the rotated 3D points onto the X-Z plane (Step 2-1). Then, extract the face polygon of the panel (Step 2-2).
- Step 3: Convert 2D polygon to 3D polygon-pair. Use the Circle-K-neighbors approximation V2 to convert the extracted 2D polygon into 3D polygon-pair.
- Step 4: Inverse rotation. Rotate the 3D polygon-pair around the Z-axis by $\theta$ to finally obtain the front and back face polygons of the panel.

*3) Light part*

Like pole objects, extracting geometric information for light objects focuses on the 2D polygon contour of a specific section. Figure 7 b) illustrates the extracting procedures, mainly including the following steps:

- Step 1: Rotate the 3D light points by $-\theta$ to parallel to the X-axis. Refer to the Panel part Step 1.

- Step 2: Chunking. First, project the light points onto the X-Z plane (Step 2-1). Then, extract the polygon and centerline of the projected points in the X-Z plane (Step 2-2). Finally, split the polygon according to the equidistantly ($\Delta l$) resampled centerline (Step 2-3). For each sub-polygon (a chunk), execute Step 3.

- Step 3: Extract the polygonal contour of the 3D points contained in the sub-polygon in the Y-Z plane. First, rotate all 3D points in the sub-polygon around the Y-axis by $-\gamma_i$ (Step 3-1), where $\gamma_i$ is the included angle between the centerline of the sub-polygon and the X-axis. Then, project the rotated points onto the Y-Z plane (Step 3-2) and extract the corresponding polygon (Step 3-3). Third, fit the smallest circle for the extracted polygon and draw $N$ rays to divide the circle equally. Collect the intersection points between the rays and the polygon to form the new polygon with $N$ vertices (Step 3-4). Finally, convert the 2D polygon to 3D using the average X-axis values of all points in the sub-polygon and rotate the 3D polygon around the Y-axis by $\gamma_i$ to offset the rotation in Step 3-1 (Step3-5).

- Step 4: Collect all 3D polygons in sequence and rotate them around the Z-axis by $\theta$.

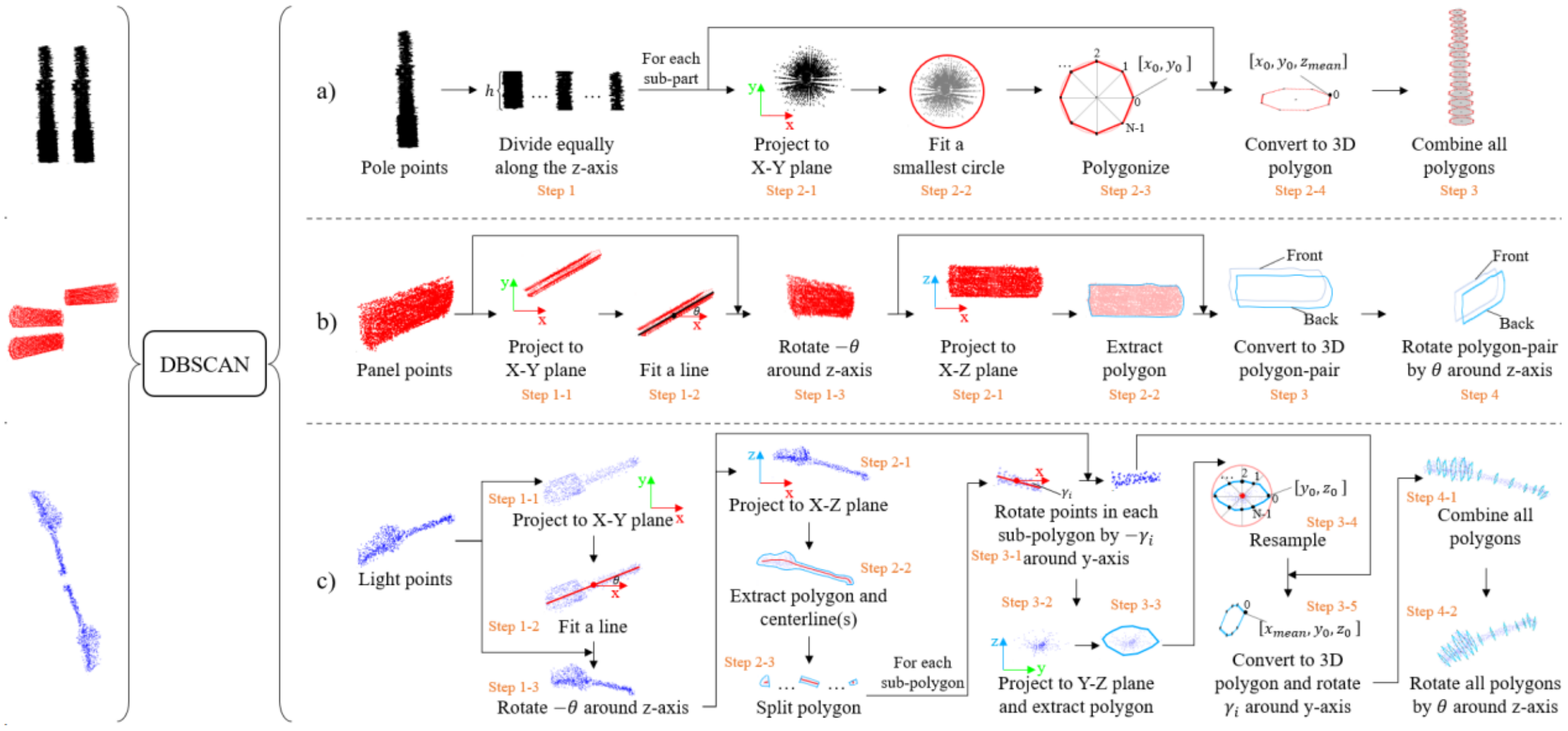

Figure 7. Geometric information extraction algorithms for Pole-like asset instances' a) pole, b) panel art, and c) light parts.

### 3.3 Geometric information formatting and storage

Existing studies rarely save the extracted geometric information but generally use a specific 3D/BIM representation method (e.g., IFC) to save the final gDTs (3D geometries) converted from the extracted geometric information. This storage strategy has some limitations. First, 3D/BIM format storage often takes up significant storage space because it needs to store all face information of the 3D geometry. Secondly, 3D/BIM format storage increases the difficulty and complexity of converting between different 3D representation forms, which requires re-extracting geometric information from the saved 3D/BIM models before converting to other formats. To reduce space occupation and improve the scalability and flexibility of data storage, transmission, and conversion, a new lightweight structured data representation method was proposed to store the extracted geometric information. Specifically, this study structured the extracted geometric information using the dictionary format of "key: value" pairs and was finally saved in the JSON files. JSON file format is simple, readable, cross-platform, and language-independent. Besides, JSON supports nested structures, which can easily represent complex data structures and relationships.

The data structures designed for different hyper-asset categories are shown in Table 1. According to Figure 5, the extracted geometric information from each Plane-like instance is multiple polygons, which are structured into a “MultiPolygon” item captaining multiple “Polygon” items. Each “Polygon” consists of a “Shell” polyline and multiple “Holes” polylines. Since the vertices of all polylines in the extracted polygons are ordered save in the vertices list, the edge information of polylines is implicitly contained in the order of the vertices, i.e., there is an edge (or line segment) between any two adjacent points in a vertices list. Therefore, the edge information is redundant, and only the “Vertices” list is finally saved in the JSON file. For Guardrail assets, each instance may be divided into one to multiple segments (as shown in Figure 6), each containing multiple polygon pairs (front and back polygons). Therefore, two “MultiPolygon” items, the “Front” and “Back”, are used to structure multiple polygon pairs. Since the front and back polygons exist in pairs, they must keep one-to-one correspondence when structuring. For Pole-like assets, each instance may contain one to multiple “Poles”, “Panels”, and “Lights”, where each pole or light is structured as a “MultiPolygon” while each panel is structured as the “Front-Back” “MultiPolygon” pairs.

Compared to storing specific 3D/BIM files, saving the extracted geometric information using the proposed data structure has some superiorities. First, the JSON file is smaller than other 3D/BIM files when representing the same 3D geometry. The proposed data structure is compact because only the essential contour information is saved instead of all the face information of the 3D geometry. Second, it is more flexible and scalable not only due to the inherent characteristics of the JOSN format but also because the extracted geometric feature information is more primitive metadata.

Table 1. Data structure for 3D geometric information storage.

| **Hyper-asset** | **Data structure** |
|---|---|
| Plane-like | { # Each plane-like instance may have one or more polygons.<br>“MultiPolygon”: { |
| | ```<br>    "Polygon_0": {<br>      "Shell": {<br>        # Ordered vertices of the 3D shell polyline. It cannot be empty.<br>        "Vertices": [[float x, float y, float z], …]<br>      },<br>      "Holes": {<br>        # Multiple 3D hole polylines. It can be empty.<br>        "Vertices": [[[float x, float y, float z], …], …]<br>      }<br>    },<br>    "Polygon_1": {…},<br>    …<br>  }<br>}<br>``` |
| Guardrail | ```<br>{  # Each guardrail instance may have one or more segments/parts.<br>  "Guardrail_0": {<br>    # Each guardrail segment is composed of the Front and Back MultiPolygon.<br>    # The front and back polygons must be in one-to-one correspondence.<br>    "Front": {<br>      "MultiPolygon": {…}<br>    },<br>    "Back": {<br>      "MultiPolygon": {…}<br>    }<br>  },<br>  "Guardrail_1": {…},<br>  …<br>}<br>``` |
| Pole-like | ```<br>{<br>  # Each Pole-like instance has at least one pole.<br>  "Poles": {<br>    "Pole_0": {<br>      "MultiPolygon": {…}<br>    },<br>    "Pole_1": {<br>      "MultiPolygon": {…}<br>    },<br>    …<br>  },<br>  # Each Pole-like instance may have one or more panels.<br>  "Panels": {<br>    "Panel_0": {<br>      "Front": {<br>        "MultiPolygon": {…}<br>      },<br>      "Back": {<br>        "MultiPolygon": {…}<br>``` |

```
            }
        },
        "Panel_1": {…},
        …
    },
    # Each Pole-like instance may have one or more lamps.
    "Lights": {
        "Light_0": {
            "MultiPolygon": {…}
        },
        "Light_1": {
            "MultiPolygon": {…}
        },
        …
    }
}
```

### 3.4 gDT creation and representation

***1) 3D geometry creation.*** JSON data stored in Section 3.3 needs further conversion to generate the corresponding 3D geometry for each PCD instance. A 3D geometry composed of multiple polygonal faces. Similar to a polygon, a polygonal face can be represented by a set of vertices and edges. The difference is that the polygonal face has an additional normal vector indicating the face's orientation. A polygonal face can be considered as a filled (excluding holes) polygon with orientation. Therefore, the process of constructing 3D geometries is the process of constructing explicit and implicit polygonal faces from the extracted/saved geometric information. There are different construct strategies for different hyper-asset instances:

- Plane-like asset. Due to the inability of LiDAR to penetrate the ground for perception, Plane-like assets only have a planar spatial structure. Therefore, a direct strategy is constructing only the upper polygonal faces for this type of hyper-asset, as shown in Figure 8 a). Alternatively, a fixed thickness for such assets can be set based on engineering experience, i.e., copy polygon $i$ and move it $\Delta h$ downwards to form a front-back polygon pair with the original polygon. Refer to the strategy for Guardrail assets to construct the 3D geometry from the polygon pair.

- Guardrail asset. Constructing 3D geometry from polygon pairs includes the construction of both explicit and implicit faces. Explicit faces are the polygonal faces represented by front and back polygons. Implicit faces are defined by the exterior and interior (holes) polyline pair(s). As shown in Figure 8 b), AC and BD are two adjacent point pairs of the exterior polyline pair. They implicitly represent a quad-face named “ACDB” or two tri-faces named “ACD” and “ADB”. The same goes for the interior polyline pair(s). The tri-face strategy was used for demonstration in this study.

- Pole-like asset. The 3D geometry construction of this kind of hyper-asset can directly refer to the strategy for polygon pairs. For the Panel part(s), the extracted/saved geometric information is typical polyline pairs. For the Pole and Light part(s), the extracted/saved geometric information is a series of ordered polylines with vertices of the same order and quantity. Therefore, any two adjacent polylines can be considered a polyline pair, then explicit and implicit faces can be constructed.

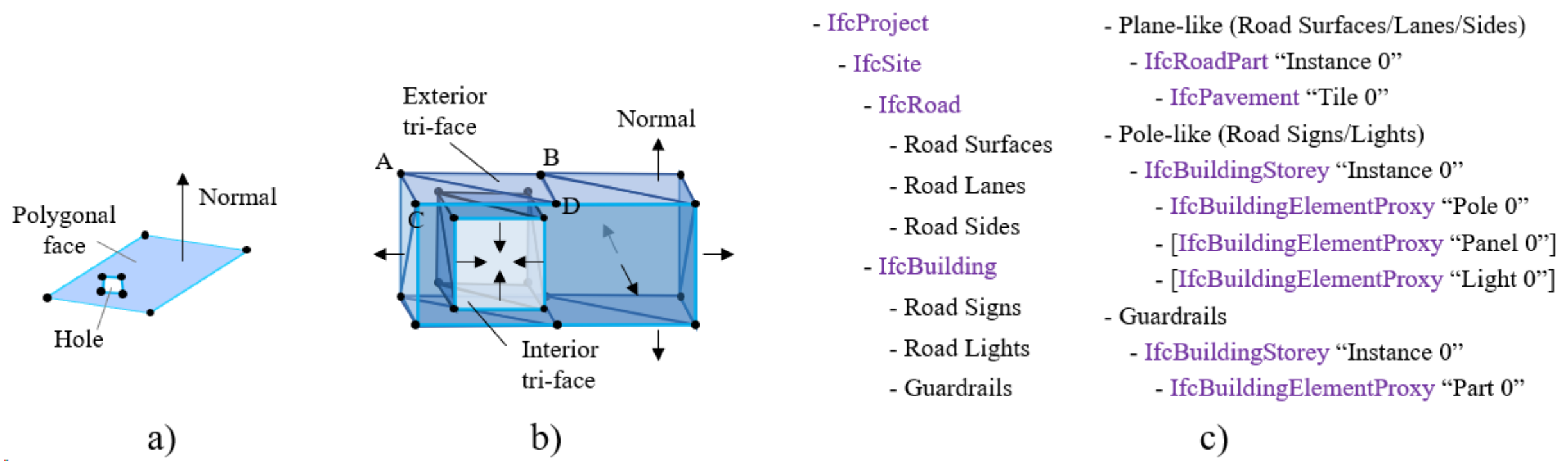

Figure 8. a) Single polygonal face with hole(s). b) 3D geometry construction from a polygon pair. c) IFC representation structure, [*] denotes possible parts.

***2) IFC representation.*** The constructed 3D geometries above are primitive/programmatic/meta descriptions, which must be further represented by a specific 3D/BIM format during field applications. The IFC format was adopted in this study as an example for conducting experiments. The boundary representation (Brep) method [9] was applied to convert the primitive gDTs to corresponding IFC entities. Figure 8 c) shows the hierarchical structure of the IFC file. The Road Surface/Lane/Side asset is represented by the IfcRoad, which may contain one or more instances defined by the IfcRoadPart. Each polygonal face of the IfcRoadPart instance is converted to an IfcPavement. The Road Sign/Light and Guardrail are represented by IfcBuilding, containing one or more instances defined by the IfcBuildingStorey. Each IfcBuildingStorey has one or more segments/parts represented by IfcBuildingElementProxy. Note that the IFC entity for each type of road asset is not unique and is only used for demonstration purposes.

## 4. Experiments

### 4.1 Experimental data

This study used six semantically labeled PCD segments of the A11 trunk road in the U.K. for experiments. The data is sourced from the *Digital Roads of the Future* project of the University of Cambridge [36]. Each PCD is approximately a 200m road segment. The original data was annotated with 12 semantic labels. However, only six types of road assets, the Road Surface, Road Lane, Road Side, Road Sign, Road Light, and Guardrail, were considered in this study based on the occurrence percentage statistics of highway objects from [11]. Figure 9 provides snapshots of six road PCD segments. For convenience, they are numbered from #1 to #6. Table 2 counts the number of instances for different asset types in each road segment (after the DBSCAN) and their total instance lengths. The length of an instance is approximated as the longest side of its oriented bounding box (OBB), i.e., $length^{inst.} = \max\,(width^{OBB}, length^{OBB}, height^{OBB})$. The total length of all instances of six road segments is up to 13,593m.

## 4.2 Experimental platform

All experiments were conducted in an HP laptop computer with an i9-13900HX CPU (2.20 GHz) and 16 GB random-access memory (RAM). The program scripts were developed based on Python 3.10 and various related packages, such as Shapely, Open3D, IfcOpenShell, etc. This study produced three types of ordered program scripts: instance segmentation, geometric information extraction (including JSON formatting and storage), and 3D geometries creation (using IFC format as an example). The outputs of the previous type of script are the only inputs for the next type of script. Therefore, each kind of script can be not only run separately but also combined as a whole to achieve the full process automation from the labeled PCD to gDT.

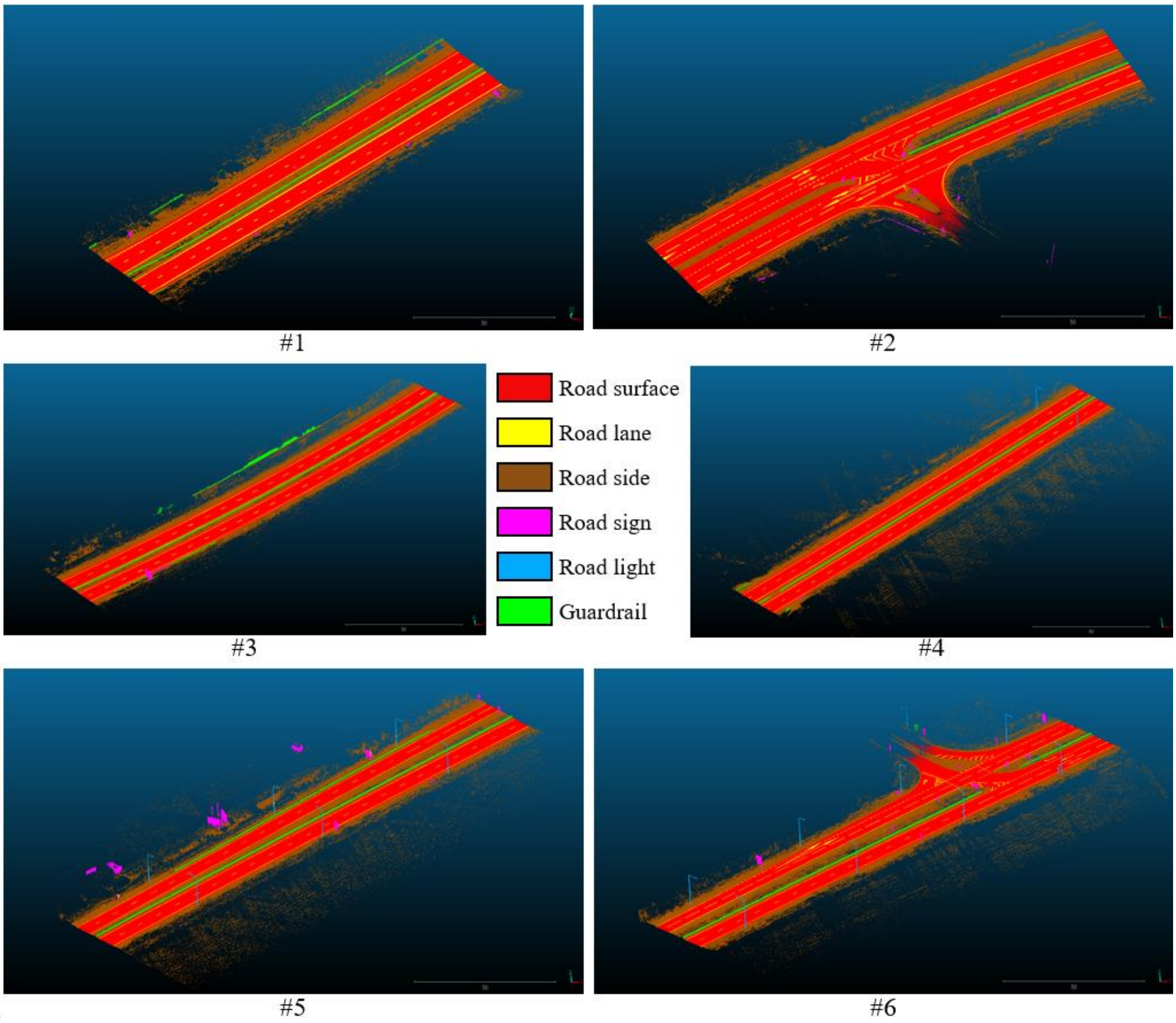

Figure 9. The six semantically labeled PCD segments for experiments.

Table 2. Detailed statistical information for each road segment.

| Road segment ID | Location | Segment length (meters) | Instance number/Total instance lengths (meters) | | | | | |
|---|---|---|---|---|---|---|---|---|
| | | | Road Surface | Road Lane | Road Side | Road Sign | Road Light | Guardrail |
| #1 | A11_570332_271173 | 200 | 2 (400)[a] | 48 (901) | 3 (600) | 4 (10) | 0 | 8 (307) |
| #2 | A11_571199_272604 | 200 | 1 (400)[b] | 143 (1056) | 6 (593) | 21 (71) | 0 | 1 (94) |
| #3 | A11_571533_272826 | 200 | 2 (400) | 50 (921) | 3 (601) | 2 (16) | 0 | 8 (322) |
| #4 | A11_571987_273209 | 200 | 2 (400) | 50 (909) | 3 (601) | 3 (7) | 2 (25) | 4 (204) |
| #5 | A11_572130_273347 | 200 | 2 (400) | 51 (935) | 3 (601) | 13 (48) | 6 (75) | 3 (328) |
| #6 | A11_572275_273485 | 200 | 1 (400)[b] | 136 (1003) | 6 (637) | 14 (42) | 9 (105) | 3 (181) |

a. (·) denotes the total length of all instances.
b. The left and right road surfaces of segment #2 are spatially connected, and they were thus considered as a single instance when conducting the DBSCAN. However, the data volume of such a single instance is equivalent to two road surface instances of other road segments. Therefore, it is considered to be two instance lengths here. The same goes for segment #5.

## 4.3 Results

### 4.3.1 IFC-based gDTs

Figure 10 a) shows some instance-level results of the extracted geometric information and IFC representation. All explicit and implicit information is illustrated in the "Geometric information" column, where implicit refers to the triangles implicitly represented by the polygon pairs. Table A1 in Appendix A provides more instance-level results with various typical shapes. The proposed method can effectively handle multiple asset instances with representative shapes on real-world roads, such as the straight/curved road surfaces/sides/lanes, the graphic character lane markings, the triangle/rectangle/circle/arrow-shaped sign panels, the long-flat/short-thick/single-branch/double-branch light heads, and guardrails with "#", "T", and "Y" structures. In addition, the diameter changes of the pole parts can also be captured. Finally, the road segment-level result can be obtained by placing all instance-level results of a road segment into one file, as shown in Figure 10 c).

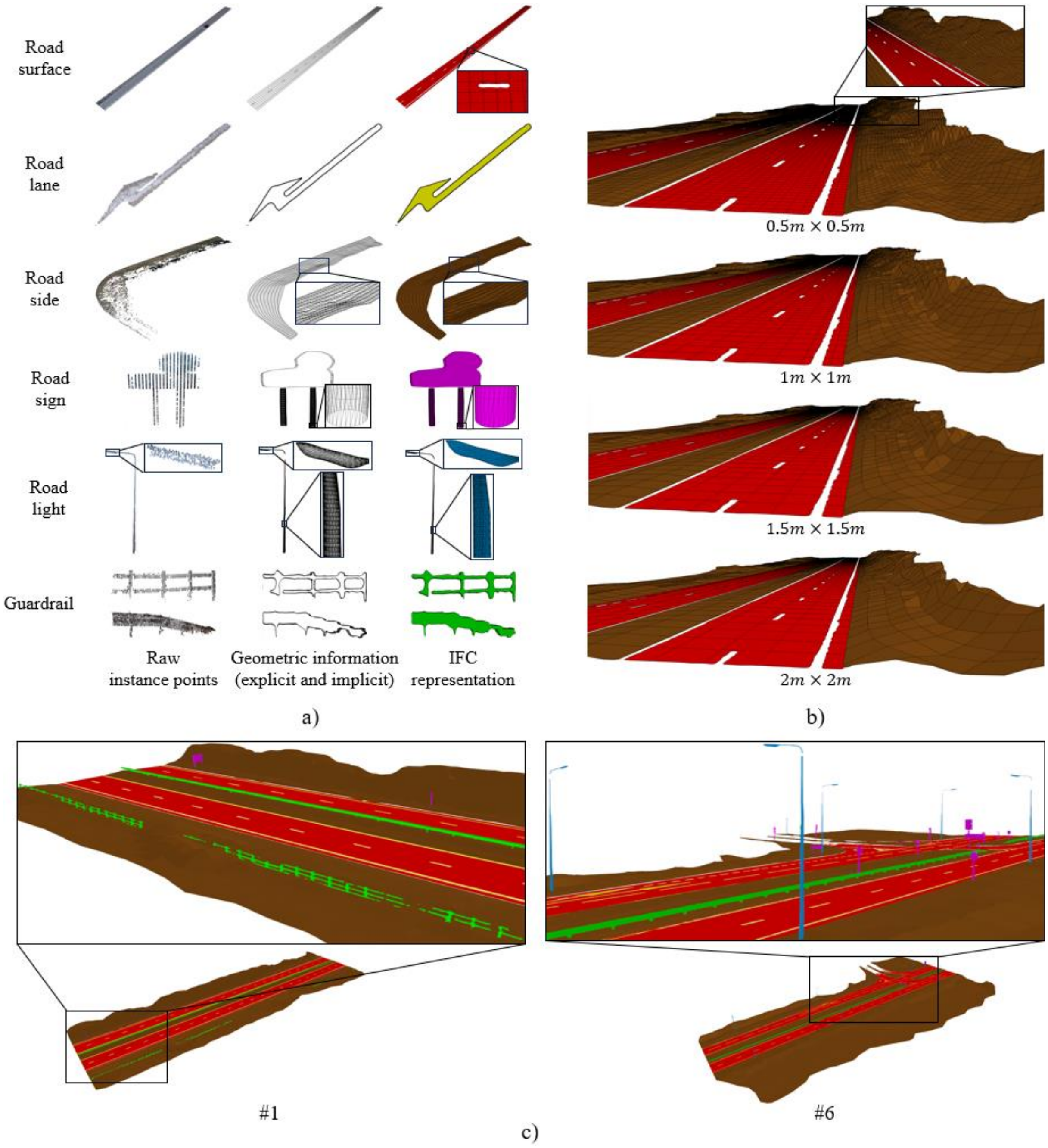

Figure 10. Geometric information extraction and IFC representation results: a) instance-level results, b) results for Road Surface and Road Side using different grid sizes, and c) road segment-level results.

For Road Surface and Road Side instances, gridding (or tiling) successfully captured the elevation fluctuations of the ground. Different grid sizes result in varying detail levels of the captured elevation fluctuations. As shown in Figure 10 b), a smaller grid size enables more detailed samplings of ground elevation, especially for Road Side instances. However, more grid-polygons are also generated, reducing the processing speed of all subsequent operations and resulting in a larger IFC file. Similarly, the developed programs also provided some key parameters to control the refinement level of the Pole/Light parts, such as the chunking size ($\Delta h$ or $\Delta l$) and polyline resample number ($N$), as shown in Figure A1 in Appendix A. Smaller $\Delta h/\Delta l$ and larger $N$ usually mean more refined results. However, a smaller $\Delta h/\Delta l$ also indicates that each chunk contains fewer valid points representing the profile, making the results of the extracted sectional polygon more sensitive to outliers and producing jagged edges.

### 4.3.2 Performance evaluation

This study adopted two quantitative metrics to comprehensively evaluate the performance of the proposed method: the accuracy of the generated gDTs and the total time from labeled points to gDTs.

***1) Accuracy.*** The unsigned/absolute distance [37] of the ground-truth (GT) points to the generated gDTs was used to measure the accuracy of the generated gDTs. An unsigned distance function $f: \mathbb{R}^3 \rightarrow \mathbb{R}$ maps 3D points $x \in \mathbb{R}^3$ to the nearest distance between $x$ and some surface $S$:

$$f(x) = \min_{x' \in S} |x - x'|$$

For any $x \in \mathbb{R}^3$, $f(x) \geq 0$. If the point $x$ lies on the surface, then $f(x) = 0$. Smaller unsigned distance means higher accuracy.

Table 3 calculates the average (Avg) and standard deviation (Std) of the unsigned distance between the GT points and generated gDTs for each asset category in one road segment. The generated gDTs of six road segments achieve an overall accuracy of 1.46cm in unsigned distance, with a standard deviation of 1.75cm. Comparing the results of each asset type, the Plane-like asset achieved the highest accuracy with an average distance of only 0.13cm and a variance of 0.4cm. The next are the Road Light, Road Sign, and Guardrail assets, achieving average distances of 1.55cm, 2.59cm, and 4.27cm, respectively. Table 4 further explores the gDT accuracy of using different grid sizes for the Road Surface/Side assets. Intuitively, the average distance and variance gradually decrease as the grid size gets smaller, i.e., smaller grid size refers to higher gDT accuracy. However, the accuracy improvement from one grid size to another also decreases as the grid size gets smaller. For example, the average distance improvement from $2m \times 2m$ to $1.5m \times 1.5m$ is 0.07cm, while the improvement is only 0.02cm from $1m \times 1m$ to $0.5m \times 0.5m$.

Table 3. The accuracy of the generated gDTs measured by the unsigned distance (in cm).

| Road segment | | Road Surface | Road Side | Road Lane | Road Sign | Road Light | Guardrail | **Mean** |
|---|---|---|---|---|---|---|---|---|
| #1 | Avg | 0.01 | 0.19 | 0.05 | 2.36 | / | 4.13 | 1.35 |
| | Std | 0.05 | 0.29 | 0.10 | 2.78 | / | 3.51 | 1.35 |
| #2 | Avg | 0.09 | 0.27 | 0.21 | 2.35 | / | 4.61 | 1.50 |
| | Std | 0.87 | 0.46 | 1.14 | 4.15 | / | 3.59 | 2.04 |
| #3 | Avg | 0.01 | 0.35 | 0.07 | 2.79 | / | 3.98 | 1.44 |
| | Std | 0.07 | 0.49 | 0.15 | 3.48 | / | 3.24 | 1.49 |
| #4 | Avg | 0.01 | 0.21 | 0.05 | 1.05 | 1.27 | 4.41 | 1.17 |
| | Std | 0.05 | 0.32 | 0.12 | 1.22 | 1.88 | 3.48 | 1.18 |
| #5 | Avg | 0.03 | 0.16 | 0.12 | 4.30 | 1.62 | 3.91 | 1.69 |
| | Std | 0.17 | 0.30 | 0.43 | 7.05 | 2.05 | 3.15 | 2.19 |
| #6 | Avg | 0.13 | 0.15 | 0.16 | 2.69 | 1.76 | 4.55 | 1.57 |
| | Std | 1.43 | 0.26 | 0.55 | 4.21 | 2.17 | 3.52 | 2.02 |
| #1-6 | Avg | 0.04 | 0.22 | 0.11 | 2.59 | 1.55 | 4.27 | **1.46** |
| | Std | 0.44 | 0.35 | 0.41 | 3.82 | 2.03 | 3.42 | **1.75** |

* The applied grid size for the Road Surface/Side is 1m×1m, while the $\Delta h/\Delta l$ and $N$ for the Pole/Light part are 0.1m and 30.

Table 4. The accuracy of the generated gDTs of the Road Surface/Side assets using different grid sizes (in cm).

| Road segment | Grid size | Road Surface | | Road Side | | **Mean** | |
|---|---|---|---|---|---|---|---|
| | | Avg | Std | Avg | Std | Avg | Std |
| #1-6 | $0.5m \times 0.5m$ | 0.04 | 0.42 | 0.19 | 0.31 | **0.11** | **0.36** |
| | $1m \times 1m$ | 0.04 | 0.44 | 0.22 | 0.35 | 0.13 | 0.40 |
| | $1.5m \times 1.5m$ | 0.08 | 0.73 | 0.28 | 0.43 | 0.18 | 0.58 |
| | $2m \times 2m$ | 0.18 | 1.19 | 0.32 | 0.47 | 0.25 | 0.83 |

***2) Efficiency.*** As a fully automated solution, the overall execution efficiency from labeled points to IFC-based gDTs is equally important as the accuracy. Table 5 calculates the time cost in the process of geometric information extraction and gDT creation & IFC representation. Overall, a total of 2162 seconds (about 36 minutes) is required to convert six road PCD segments to IFC-based gDTs, with an average of 360 seconds (6 minutes) per segment (200m). If viewed from the perspective of total instance length, the efficiency of the developed method reaches 6.29 m/s. Besides, the vast majority of the time (over 92%) is used for geometric information extraction, with the Road Surface/Side and Guardrail assets being the most time-consuming. One of the main reasons is that the data volume and spatial span of these types of asset instances are relatively large. Unfortunately, the Alphashape's efficiency in extracting polygonal contours decreases with increasing data volume. Additional statistics found that extracting polygonal contours accounted for over 90% of the total time consumed in geometric information extraction.

Table 6 further counts the time cost for the Road Surface/Side assets using different grid sizes. The smaller the grid size, the more time is required as a larger data volume (more tiles) needs to be processed by the programs. Note that regardless of the grid size, the Alphashape algorithm for extracting polygonal contours costs approximately the same time. Therefore, the extra time cost from $2m \times 2m$ to $0.5m \times 0.5m$ is mainly consumed by processing the newly added sampling points/tiles. However, comparing the results in Table 4 and Table 6, the accuracy only improves by 0.02cm from $1m \times 1m$ to $0.5m \times 0.5m$, but requires an additional execution time of 382 seconds. Therefore, using $1m \times 1m$ or $1.5m \times 1.5m$ might be the optimal choices to achieve ideal performance in both accuracy and efficiency.

Table 5. Time cost (in seconds) of geometric information extraction and gDT creation & IFC representation.

| Asset type | Geometric information extraction | | | | | | gDT creation & IFC representation | | | | | | **Total** |
|---|---|---|---|---|---|---|---|---|---|---|---|---|---|
| | #1 | #2 | #3 | #4 | #5 | #6 | #1 | #2 | #3 | #4 | #5 | #6 | |
| Road Surface | 144 | 218 | 158 | 155 | 143 | 188 | 3 | 6 | 3 | 3 | 3 | 5 | 1029 |
| Road Side | 54 | 50 | 46 | 55 | 63 | 64 | 4 | 4 | 4 | 10 | 8 | 8 | 370 |
| Road Lane | 22 | 27 | 21 | 20 | 15 | 21 | 1 | 2 | 1 | 1 | 1 | 1 | 133 |
| Road Sign | 3 | 9 | 11 | 1 | 30 | 24 | 1 | 5 | 3 | 1 | 13 | 6 | 107 |
| Road Light | / | / | / | 5 | 15 | 22 | / | / | / | 3 | 10 | 14 | 69 |
| Guardrail | 63 | 26 | 106 | 69 | 108 | 47 | 8 | 2 | 8 | 5 | 8 | 4 | 454 |
| **Total** | 2003 | | | | | | 159 | | | | | | **2162** |

* The applied grid size for the Road Surface/Side is 1m×1m, while the Δh/Δl and $N$ for the Pole/Light part are 0.1m and 30.

Table 6. Time cost (in seconds) of geometric information extraction and gDT creation & IFC representation for the Road Surface/Side asset using different grid sizes.

| Road segment | Grid size | Geometric information extraction | | | gDT creation & IFC representation | | | **Total** |
|---|---|---|---|---|---|---|---|---|
| | | Road Surface | Road Side | **Total** | Road Surface | Road Side | **Total** | |
| #1-6 | $0.5m \times 0.5m$ | 1190 | 378 | 1568 | 63 | 149 | 212 | 1780 |
| | $1m \times 1m$ | 1006 | 332 | 1338 | 22 | 38 | 60 | 1398 |
| | $1.5m \times 1.5m$ | 929 | 309 | 1238 | 13 | 18 | 31 | 1269 |
| | $2m \times 2m$ | 908 | 298 | 1206 | 10 | 10 | 20 | 1226 |

### 4.3.3 Comparison of different storage strategies

Additional comparisons were made between the two storage strategies: 1) saving the extracted geometric information in JSON files using the proposed data structure and 2) saving the final converted IFC files. As shown in Figure 11, both files have a certain level of readability since they all use text for information storage. However, JSON files are more concise and readable, as their "key: value" dictionary-based nested structure can directly provide hierarchical relationships between road components or functional parts. Although IFC files use line numbers to mark relationships between different entities implicitly, multiple full-text retrievals are required to determine a complete chain of relationships. Besides, storing the extracted geometric information in JSON takes up less disk space than storing the final converted IFC files. As shown in Table 7, storing six road segments' geometric information in JSON format requires only 48.1 Megabytes (MB) of disk space, while the converted IFC files are about five times that of JSON files, reaching 236.71MB. Finally, as a prevailing lightweight data exchange format, JSON features platform independence, multiple programming language support, ease of parsing and generation, and efficient Internet transmission. Therefore, the JSON format is more suitable for data exchange between clients and servers than IFC in the era of the IoT. For example, in a typical future application scenario, when the client is viewing gDTs of a specific road segment/asset, the server only needs to transmit the corresponding JSON file to the client. The client then converts the received JSON file into an IFC file and caches it for local visualization.

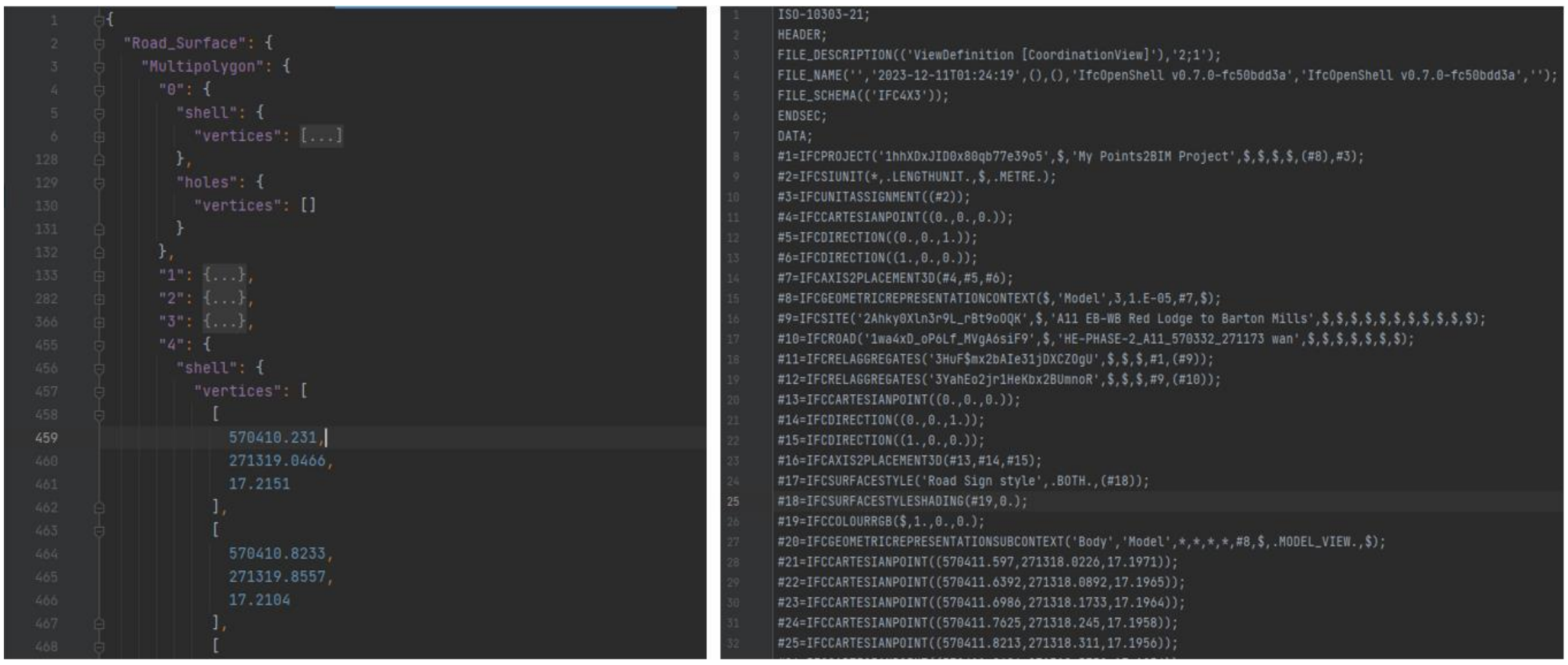

a) b)

Figure 11. Different data storage strategies: a) geometric information stored in JSON and b) 3D gDTs stored in IFC.

Table 7. File size (Megabytes, MB) of different permanent storage strategies.

| Asset type | Geometric information stored in JSON | | | | | | 3D gDTs stored in IFC | | | | | |
|---|---|---|---|---|---|---|---|---|---|---|---|---|
| | #1 | #2 | #3 | #4 | #5 | #6 | #1 | #2 | #3 | #4 | #5 | #6 |
| Road Surface | 1.66 | 3.53 | 2.06 | 1.94 | 2.0 | 2.84 | 4.02 | 8.74 | 4.87 | 4.55 | 4.74 | 6.80 |
| Road Side | 1.83 | 1.93 | 1.73 | 4.48 | 3.70 | 3.76 | 5.04 | 5.33 | 4.76 | 12.7 | 10.4 | 10.6 |
| Road Lane | 0.66 | 1.09 | 0.89 | 0.74 | 0.76 | 1.0 | 1.24 | 2.12 | 1.67 | 1.39 | 1.42 | 1.94 |
| Road Sign | 0.17 | 0.57 | 0.31 | 0.06 | 1.33 | 0.64 | 1.60 | 8.83 | 4.66 | 0.95 | 20.3 | 9.83 |
| Road Light | / | / | / | 0.32 | 0.93 | 1.36 | / | / | / | 4.85 | 14.7 | 21.6 |
| Guardrail | 1.46 | 0.34 | 1.21 | 0.83 | 1.30 | 0.67 | 12.4 | 3.46 | 12.5 | 8.53 | 13.4 | 6.77 |
| **Total** | **48.1** | | | | | | 236.71 | | | | | |

* The applied grid size for the Road Surface/Side is 1m×1m, while the Δh/Δl and *N* for the Pole/Light part are 0.1m and 30.

## 5. Discussion

This study proposes and has successfully implemented a novel scan-to-BIM solution for twinning as-built roads from labeled scanning points. With the following advantages and characteristics, the developed solution has great practical prospects and is of great significance for the low-cost and automated digital twinning of as-built roads:

- More accurate. First of all, the proposed method is entirely based on the LiDAR scanning data, which inherently has higher accuracy in objects' 3D size perception than methods based on single or multi-view images. High-precision raw PCD lays a solid data foundation for creating more accurate road gDTs. Second, according to experiments conducted in this study on six real-world road segments, the average distance from the GT points to the generated gDTs is as small as 1.46cm, with a standard deviation of 1.75cm. This number for Plane-like assets is even smaller, only 0.13cm (Table 3), reaching the millimeter-level accuracy. Compared to the latest related research, Lu et al. [9] achieved a distance error of 7.05cm in bridge twinning based on labeled PCD, while Jiang et al. [16] got a distance error of 15.76cm in road twinning based on map data.

- More efficient. The proposed framework is fully automatic, from labeled points to the final IFC-based gDTs. Each algorithm described in Section 3 was packaged as a callable application programming interface (API). Therefore, these three types of APIs can be called not only separately to meet specific requirements but also sequentially in a new script to achieve automation from points to gDTs. Besides, the proposed method achieves a twinning speed of 6 minutes per 200m segment. Or, from the perspective of total instance length, the speed reaches 6.29m/s. Both these two speeds are significantly better than manual creation efficiency (Table 1 in [11]).

- Shape-independent. The proposed method is applicable for twinning various-shaped road assets by unifying the geometric information extraction of diverse road assets into the extraction of sectional polygon contours. For Plane-like assets, their geometric information mainly refers to the contour polygon in a specific 2D plane. For other 3D assets (e.g., road signs/lights and guardrails), their contours in a particular plane or section can be obtained through projection or chunking, which are then combined to form a polygon pair or series to get the final geometric information. The most crucial of the above extraction algorithms is the Alphashape algorithm used for contour polygon extraction, which is shape-independent and applicable to arbitrary shapes. Therefore, the proposed solution can also be regarded as shape-independent in theory.

- Flexible and controllable. In practical demand scenarios, there might be different precision-level and time requirements for twinning various types of roads with diverse conditions. The proposed solution provides rich parameters for users to customize the refinement and size of the gDTs generated, flexibly meeting the demands of various refinement-level and time requirements.

- Highly scalable. Instead of storing the final 3D files, this study uses a novel data structure to store the extracted geometric information in JSON files. In addition, the created 3D geometries from the extracted/stored geometric information are in primitive representation, i.e., multiple polygonal faces composed of points and lines. Therefore, the scalability is first reflected in the fact that these primitive representations provide an easy and native way to generate various 3D/BIM file formats to meet the native support for multiple 3D file formats in the future. On the other hand, the high scalability also benefits from the nested structure of JSON format, which allows managers/developers to easily adjust the data structure of geometric information by adding or deleting "key: value" pairs according to future demands. JSON format also features platform independence and multiple programming language support.

## 6. Conclusion

A novel and practical scan-to-BIM framework is proposed in this study for automated twinning as-built roads from semantically labeled PCD, which focuses on geometric information extraction & structured storage and gDT creation & representation. The proposed method identifies the six considered road assets into Plane-like (Road Surface/Side/Lane), Pole-like (Road Sign/Light), and Guardrail hyper-asset instances and further segments the Pole-like instances into multiple functional parts. Targeted algorithms/solutions are developed for each hyper-asset instance or functional part from processes of geometric information extraction to gDT representation. The experimental results on six real-world road PCD segments show that the proposed method achieves state-of-the-art performance in both accuracy and efficiency in twinning as-built roads, with an average distance error of 1.46cm and a twinning speed of 6 minutes per 200 meters (or 6.29 m/s based on the total instance length). The implemented scan-to-BIM framework has the characteristics of high accuracy, efficiency, applicability, flexibility, and scalability, which is expected to save considerable labor, time, and money for twinning the as-built roads.

However, there are still some limitations. First, this study mainly focuses on geometric information extraction and 3D geometry creation & representation from labeled PCD, with less attention paid to the semantic/instance/part segmentation and related model optimization experiments. In fact, the proposed solution heavily depends on the accuracy of those segmentation results. Therefore, future work can start with constructing a more accurate PCD for each asset instance/part, such as applying more advanced deep learning-based models for semantic/instance/part segmentation to obtain more accurate PCD instances. In addition, the proposed method lacks solutions to handle occlusion issues. For example, the roadside vegetation may cause occlusions on Pole-like and Guardrail assets, resulting in patchy PCD and thus affecting the quality of the generated gDTs. Therefore, PCD completion models can also be studied and applied in future studies to repair/complete the occluded parts for each instance. Finally, there is significant room for improvement in generating gDTs for Guardrail assets. Follow-up studies can refer to Pole-like assets to further divide guardrails into multiple functional parts, such as the post/support and rail, and then design targeted algorithms to process each part.

**Acknowledgment**

The Shenzhen Science and Technology Innovation Committee Grant #JCYJ20180507181647320 and General Research Fund from Research Grant Council of Hong Kong SAR #11211622 jointly supported this work. The conclusions herein are those of the authors and do not necessarily reflect the views of the sponsoring agencies.

## Appendix A

Table A1. More examples of instance-level results.

| Instance type | Raw instance points | Geometric information (explicit and implicit) | IFC representation |
|---|---|---|---|
| Road Surface | 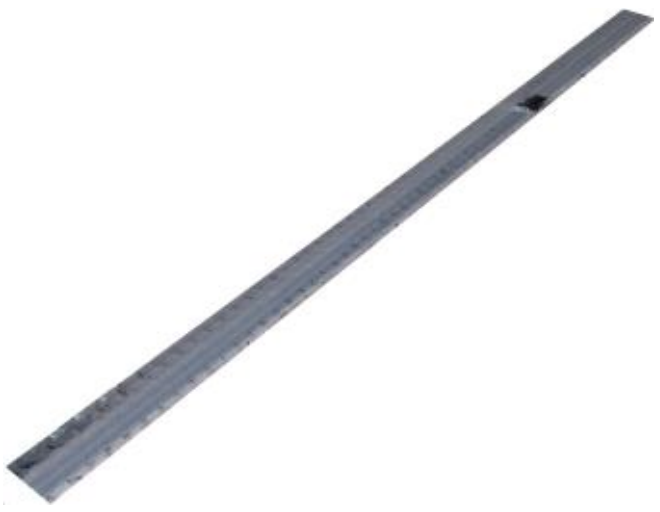 | 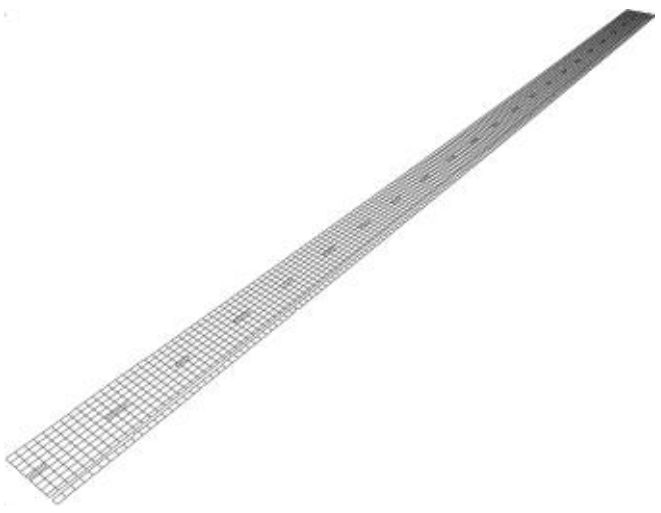 | 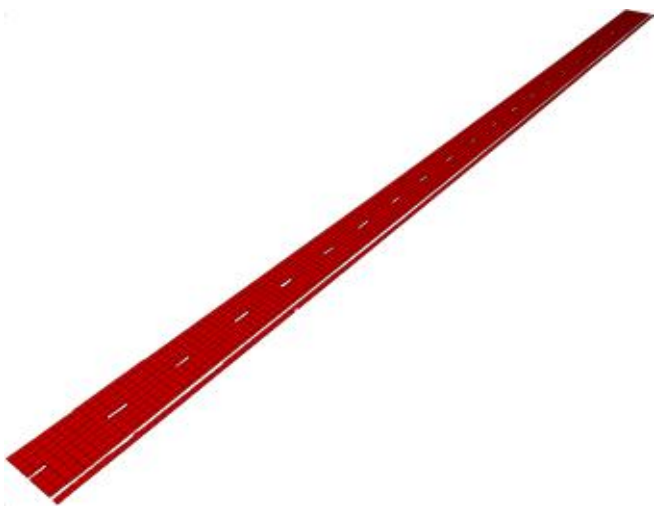 |
| | 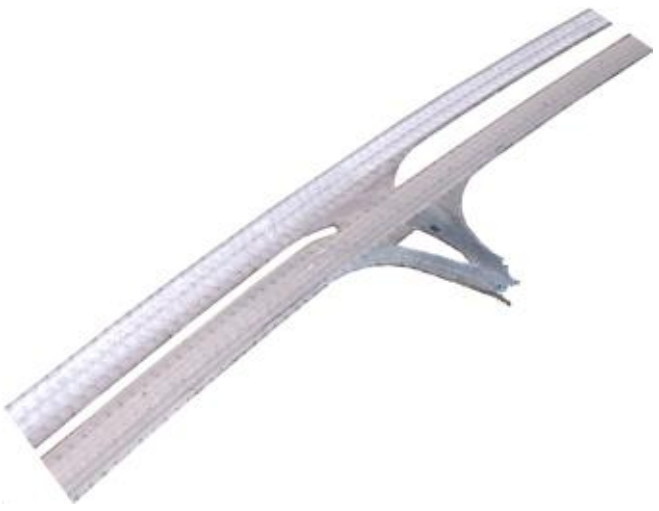 | 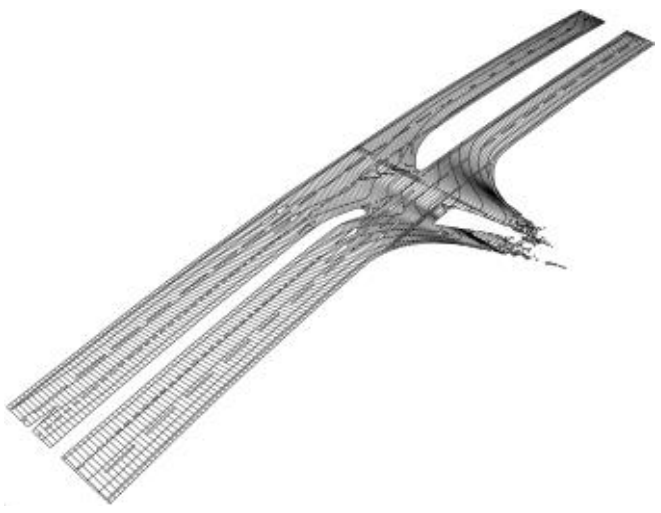 | 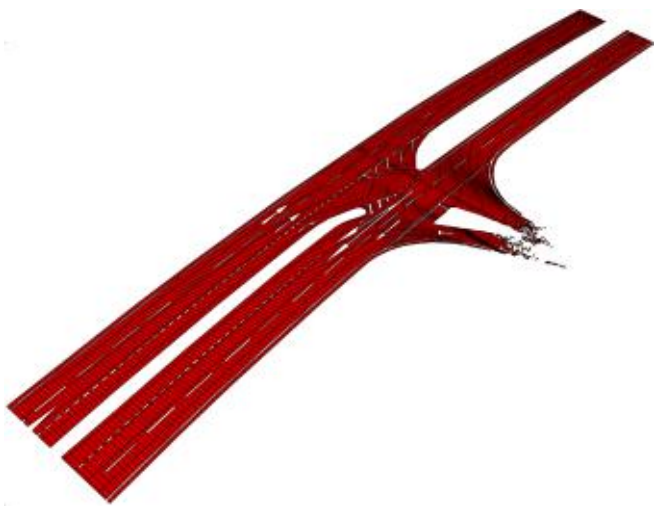 |

| | |
|---|---|
| Road Side | 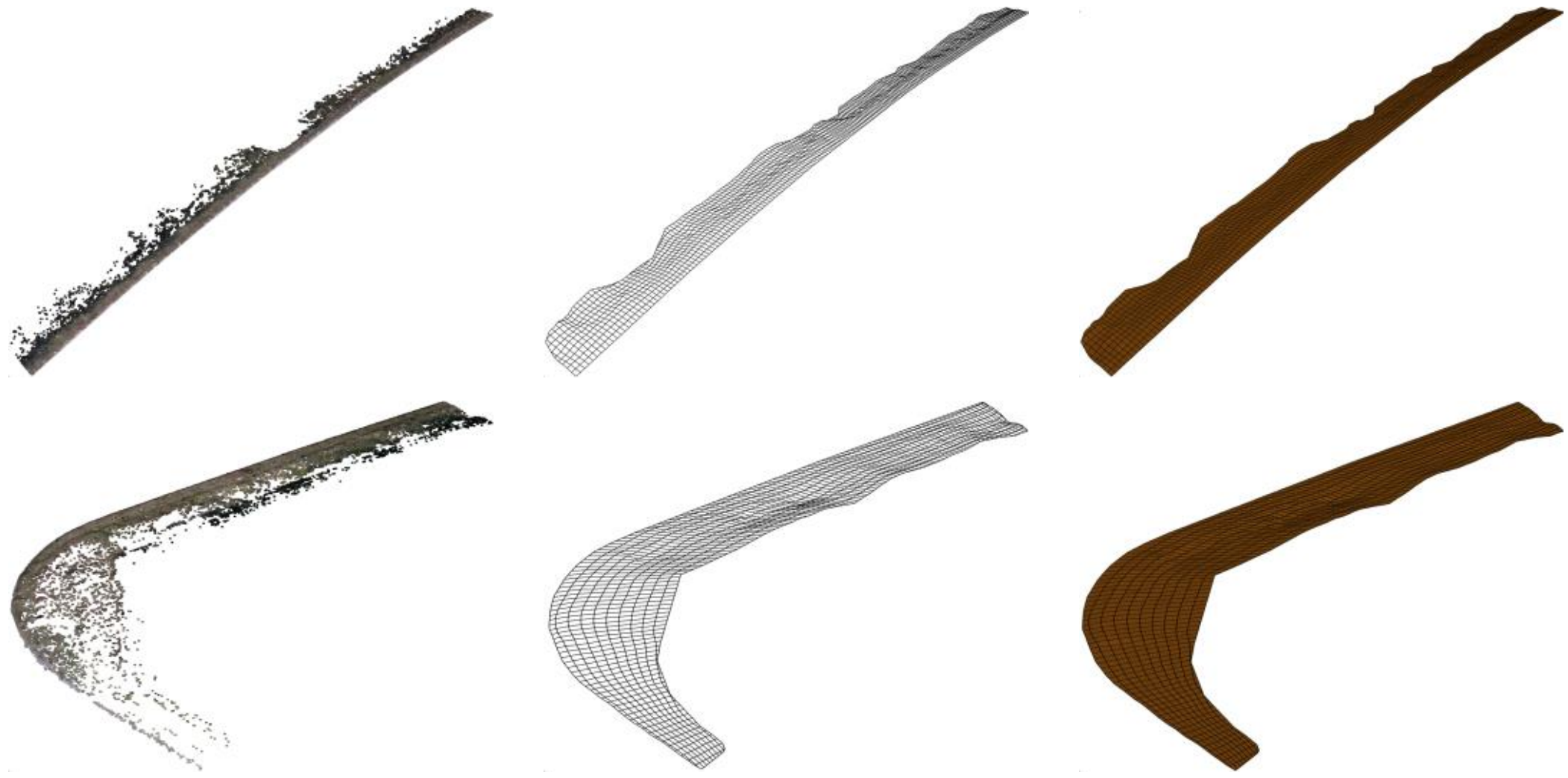 |
| Road Lane | 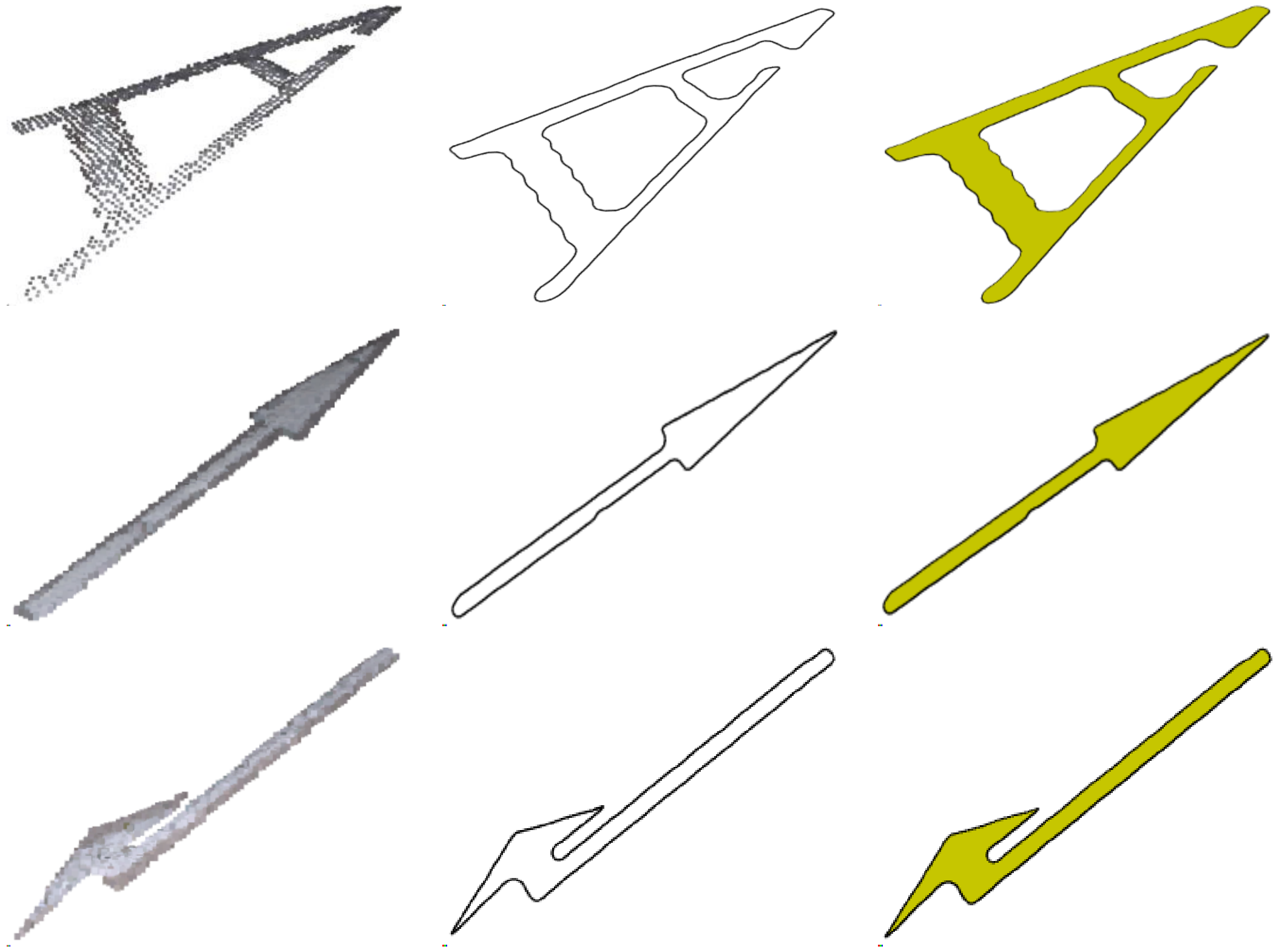 |
| Road Sign | 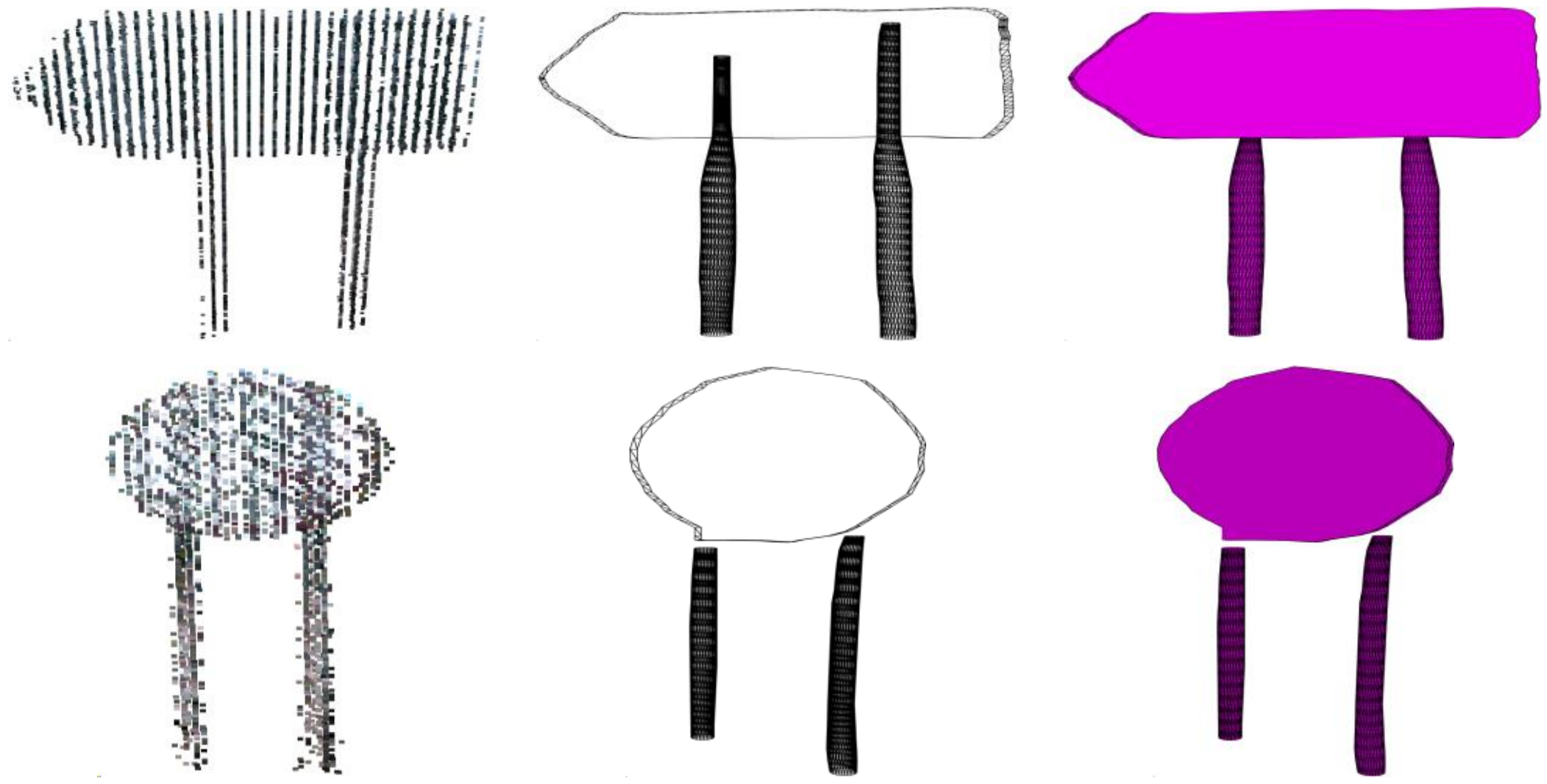 |

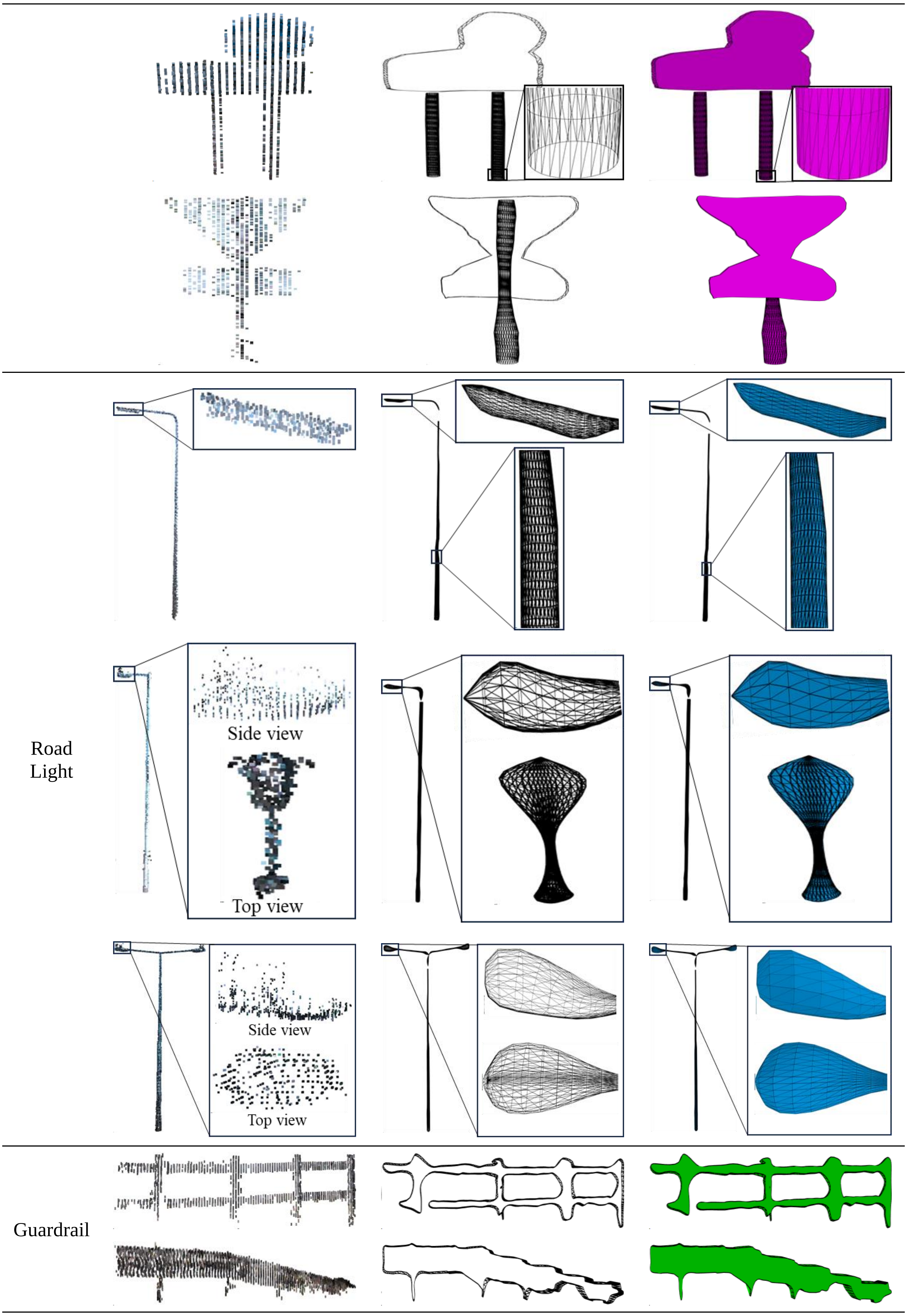

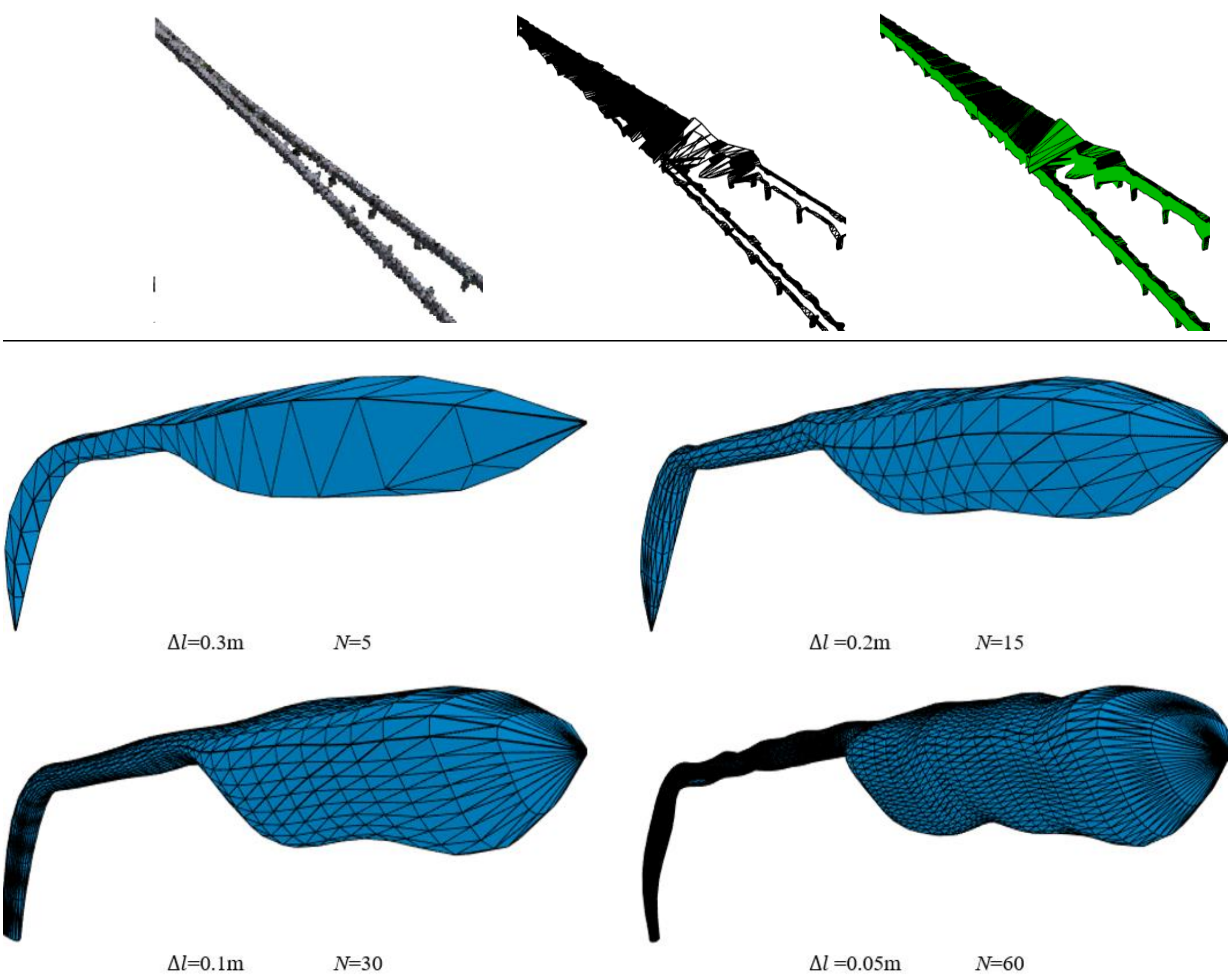

Figure A1. The created gDT of the Light part using different parameter settings.